\documentclass[preprint,12pt]{elsarticle}

\usepackage{amssymb}
\usepackage{xcolor}
\usepackage{algorithm}
\usepackage{amsmath}
\usepackage{algcompatible}
\usepackage{algpseudocode}
\usepackage{amsthm}
\usepackage{booktabs}
\usepackage{multirow}
\usepackage{graphicx}
\usepackage{tikz}
\usepackage{pgfplots}
\pgfplotsset{compat=1.18}
\usepackage{subfig}
\usepackage{url}
\graphicspath{{./}}

\newcounter{bla}

\journal{Computer Physics Communications}

\begin{document}

\begin{frontmatter}


\title{RL-ABC: Reinforcement Learning for Accelerator Beamline Control}

\author[a]{Anwar Ibrahim\corref{author}}
\author[a]{Fedor Ratnikov}
\author[a]{Maxim Kaledin}
\author[b]{Alexey Petrenko}
\author[a]{Denis Derkach}

\cortext[author]{Corresponding author.\\\textit{E-mail address:} aibrahim@hse.ru}
\address[a]{Laboratory of Methods for Big Data Analysis, HSE University, Moscow, Russia}
\address[b]{Budker Institute of Nuclear Physics, Novosibirsk, Russia}

\begin{abstract}
Particle accelerator beamline optimization is a high-dimensional control problem traditionally requiring significant expert intervention. We present RLABC (Reinforcement Learning for Accelerator Beamline Control), an open-source Python framework that automatically transforms standard Elegant beamline configurations into reinforcement learning environments. RLABC integrates with the widely-used Elegant beam dynamics simulation code via SDDS-based interfaces, enabling researchers to apply modern RL algorithms to beamline optimization with minimal RL-specific development.

The main contribution is a general methodology for formulating beamline tuning {as a finite-horizon decision process}: RLABC automatically preprocesses lattice files to insert diagnostic watch points before each tunable element, constructs a 57-dimensional {observation} from beam statistics, {correlation information}, and aperture constraints, and provides a configurable reward function for transmission optimization. {The environment follows the Gymnasium interface and can therefore be coupled to continuous-control RL algorithms; the present study evaluates DDPG and implements an optional stage-learning curriculum.}

{Validation on a test beamline derived from the VEPP-5 injection complex (37 control parameters across 11 quadrupoles and 4 dipoles) demonstrates that the framework successfully enables RL-based optimization, with a Deep Deterministic Policy Gradient agent achieving 70\% particle transmission.} The framework's stage learning capability allows decomposition of complex optimization problems into manageable subproblems { and provides an optional warm start for later training stages.} The complete framework, including configuration files and example notebooks, is available as open-source software to facilitate adoption and further research.
\end{abstract}


\end{frontmatter}

\noindent \textbf{PROGRAM SUMMARY}

\begin{small}
\noindent
{\em Program Title:} RLABC (Reinforcement Learning for Accelerator Beamline Control) \\

{\em CPC Library link to program files:} (to be added by Technical Editor) \\

{\em Developer's repository link:} \texttt{https://github.com/Anwar9Ibrahim/RL-ABC.git} \\

{\em Licensing provisions:} MIT \\

{\em Programming language:} Python 3.10 \\

{\em Supplementary material:} Example lattice and simulation files for the Elegant beam dynamics code (.lte, .ele), configuration files, pretrained model checkpoints, and evaluation logs \\

{\em Nature of problem:} \\
Applying reinforcement learning to particle accelerator beamline optimization requires addressing several interrelated challenges. {Beamline tuning is naturally a feedback control process, and casting it as reinforcement learning requires a careful Markov decision process formulation. Agent observations must summarize relevant beam physics in fixed-dimensional vectors suitable for neural networks.} The system must integrate with existing simulation infrastructure, and researchers need the flexibility to experiment with different algorithms and configurations. Existing solutions require significant custom development for each beamline, limiting adoption of RL techniques in the accelerator physics community.

{\em Solution method:} \\
RLABC provides an automated pipeline for transforming Elegant beamline configurations into RL environments. Given standard Elegant lattice files (\texttt{.lte}) and command files (\texttt{.ele}), RLABC parses and preprocesses the beamline to insert diagnostic watch points before each tunable element, constructs a graph representation for efficient queries about element relationships, extracts a fixed 57-dimensional {observation} vector from beam statistics, {a correlation matrix}, and aperture parameters, and provides a configurable reward function based on particle transmission. The environment implements the Gymnasium interface (the maintained successor to OpenAI Gym) for compatibility with standard RL libraries (Stable-Baselines3). Stage learning strategies enable curriculum-based training that progressively increases problem complexity.

{\em Additional comments including restrictions and unusual features:} \\

RLABC is designed as a flexible research platform rather than a fixed optimization tool. The main features are: automatic adaptation to different beamline layouts; modular architecture allowing experimentation with alternative {observation representations}, reward functions, and RL algorithms; integration with Elegant via SDDS data exchange; and stage learning support { as an optional curriculum mechanism}. The framework requires a working installation of the Elegant simulation program~\cite{borland2000} and its SDDS Toolkit as external prerequisites. The framework emphasizes training in simulation prior to potential deployment on real hardware. Computational cost is dominated by beam dynamics simulation; {measured single-simulation and complete-episode timings are reported in Section~\ref{sec:test_setup}.} Future integration with accelerated simulators could substantially reduce training times. The repository includes configuration files, pretrained model checkpoints, and documentation intended for extension by both accelerator physicists and RL researchers.

\end{small}

\section{Introduction}
\label{sec:intro}

Particle accelerators are essential scientific instruments playing a central role in fundamental physics, materials science, medical applications, and industrial processing~\cite{lee2018,wiedemann2015,chao2013,wolski2014}. Despite their diversity, all particle accelerators share common subsystems, including beam sources, transport and injection systems, accelerating structures, and diagnostic and control systems.

This work focuses on the \textit{injection complex}, or \textit{beamline}, responsible for transporting the particle beam from the source to the main accelerator. A typical beamline comprises particle sources, RF acceleration cavities, dipole magnets for beam steering, quadrupole magnets for transverse focusing, diagnostic systems (beam position monitors, profile monitors), and vacuum infrastructure. The beamline tuning problem involves optimizing magnet parameters to maximize particle transmission while maintaining beam quality.

To illustrate, a test beamline is considered in this work, a positron transport segment from the VEPP-5 injection complex, that contains 11 quadrupole magnets (each with strength parameter $K_1$ and optional dipole correction kicks HKICK and VKICK) and 4 dipole magnets (with fractional strength error FSE), yielding a 37-dimensional continuous parameter space. The strong coupling between these parameters, combined with nonlinear beam dynamics and particle losses at apertures, can make traditional optimization approaches time-consuming and may lead to suboptimal results.

Traditionally, beamline tuning is performed by expert operators leveraging deep accelerator physics knowledge, or through mathematical approaches such as simplex algorithms~\cite{nelder1965} or Bayesian Optimization~\cite{pelikan2005bayesian}. However, simplex algorithms are generally inefficient for noisy, multi-dimensional, coupled optimization problems, while Bayesian optimization, though sample-efficient, may struggle in dynamic environments and can become trapped in local optima. \textit{Reinforcement Learning (RL)} offers an alternative paradigm for sequential control problems, with the potential to learn policies that handle high-dimensional, coupled parameter spaces~\cite{sutton2018}.

RL is successfully applied to various domains including robotics~\cite{kober2013,tang2025deep}, autonomous driving~\cite{kendall2018learning}, game playing~\cite{mnih2015}, and real-world decision-making~\cite{nambiar2023deep,bai2025review}. Its ability to handle high-dimensional spaces, temporal dependencies, and nonlinear system dynamics makes it well-suited for accelerator beamline control.

Despite this potential, applying RL to beamline optimization is far from straightforward. {Beamline tuning is a feedback process, and casting it as RL requires explicit definitions of the state, observation, action, transition, and reward. The agent observation must summarize the beam information used by the policy while remaining fixed in size. The whole system needs to interface cleanly with existing simulation tools, and researchers need the flexibility to swap algorithms and configurations without rebuilding the environment from scratch.}

We introduce \textbf{RLABC}, \textit{Reinforcement Learning for Accelerator Beamline Control}, an open-source Python framework that automates the transformation of {supported Elegant} beamline configurations into RL environments. RLABC interfaces with \textbf{Elegant}, a widely used simulation code for accelerator modeling~\cite{borland2000}, through SDDS (Self Describing Data Sets) files, providing a practical, extensible framework that enables the accelerator physics community to explore RL-based approaches with minimal overhead.

In this paper, we present the design and implementation of RLABC, describe the methodology for automatic environment construction, and validate the framework on a test beamline derived from the \textbf{VEPP-5 injection complex}~\cite{Emanov2023-wx} - an electron--positron injector complex at the Budker Institute of Nuclear Physics (BINP) in Novosibirsk, Russia. We demonstrate that RLABC successfully enables RL-based optimization, achieving results comparable to established methods.

The remainder of this paper is organized as follows. Section~\ref{sec:review} reviews state-of-the-art approaches for similar tasks. Section~\ref{sec:Theory} provides theoretical background on RL and the Elegant simulation program. Section~\ref{sec:methodology} describes RLABC's structure and implementation. Section~\ref{sec:results} presents experimental results, and Section~\ref{sec:conclusion} concludes with discussion of future work.

\section{Literature Review}
\label{sec:review}
Recent developments in machine learning open new avenues for automating some manual accelerator tuning tasks. Bayesian optimization emerges as a widely adopted framework due to its sample efficiency and robustness to noisy measurements, being applied to beamline tuning, experimental design, and control parameter optimization~\cite{rousseletal2024}. Morita \textit{et al.}~\cite{morita2023} propose combining autoencoders with Bayesian optimization to handle high-dimensional tuning problems, improving tuning speed compared to traditional Bayesian techniques alone.
{Reinforcement learning has also become an active approach for accelerator control problems in which each decision changes the conditions seen by later decisions. Demonstrations on operating machines include sample-efficient model-free RL for trajectory steering at the CERN AWAKE electron line and LINAC4~\cite{Kain2020}, surrogate-trained control of the Fermilab Booster gradient-magnet power supply with FPGA deployment~\cite{StJohn2021}, and hardware-accelerated feedback at KARA with microsecond-scale inference~\cite{scomparin2024microsecond}. Other studies address specialized tuning tasks, including superconducting radio-frequency gun optimization using a neural-network surrogate~\cite{meier2022srf} and zero-shot trend-based Soft-actor Critic (SAC) control for superconducting linac orbit and transmission tuning~\cite{chen2023trend}. Simulation-to-real studies address the limited availability and risk of real-machine training: Velotti \textit{et al.} compare explicit beam parameters with autoencoded screen-image representations and use a variational-autoencoder surrogate for the 18~MeV AWAKE electron beamline~\cite{velotti2023}, while Awal \textit{et al.} train a SAC policy with domain randomization and evaluate it in live COSY injection optimization~\cite{awal2025}. Kaiser \textit{et al.} further compare reinforcement-learning-trained optimizers with Bayesian optimization for online tuning~\cite{kaiser2024}. Together, these works show that RL can be useful for trajectory correction, injection, beam transport, and fast feedback, but they also underline recurring practical limits: sample efficiency, simulator fidelity, transfer to real hardware, and the need to define task-specific observations, actions, and rewards. \\
A related software challenge is connecting accelerator simulations to modern learning tools without rewriting the physics model for every study. Cheetah provides a PyTorch-based differentiable beam-dynamics simulator for fast machine-learning workflows~\cite{kaiser2024bridging}. RLABC takes a complementary route: it keeps Elegant as the tracking backend and automates the construction of Gymnasium-compatible RL environments from supported \texttt{.lte} and \texttt{.ele} inputs. In this sense, the contribution of RLABC is not a new RL algorithm and not a claim of universal superiority over Bayesian optimization~\cite{rousseletal2024}; it is a reusable environment-construction layer that inserts diagnostics, builds a fixed watch-point observation, exposes a unified per-element action interface, and supports stage learning for difficult beamline-tuning tasks. The simulation-based formulation underlying this framework is introduced in our earlier work~\cite{Ibrahim2025PPN,ibrahim2025IJMP}.}
\section{Theoretical Background}
\label{sec:Theory}

\subsection{Reinforcement Learning}
\label{sec:RL}

Reinforcement learning addresses goal-directed behavior through trial-and-error interaction with a dynamic environment~\cite{sutton2018}. In contexts like accelerator beamline control, RL is well-suited because tuning parameters forms a sequential process where each adjustment affects future beam states, and outcomes depend on the interplay of many coupled parameters.

\begin{figure}[!ht]
    \centering
    \includegraphics[width=0.8\textwidth]{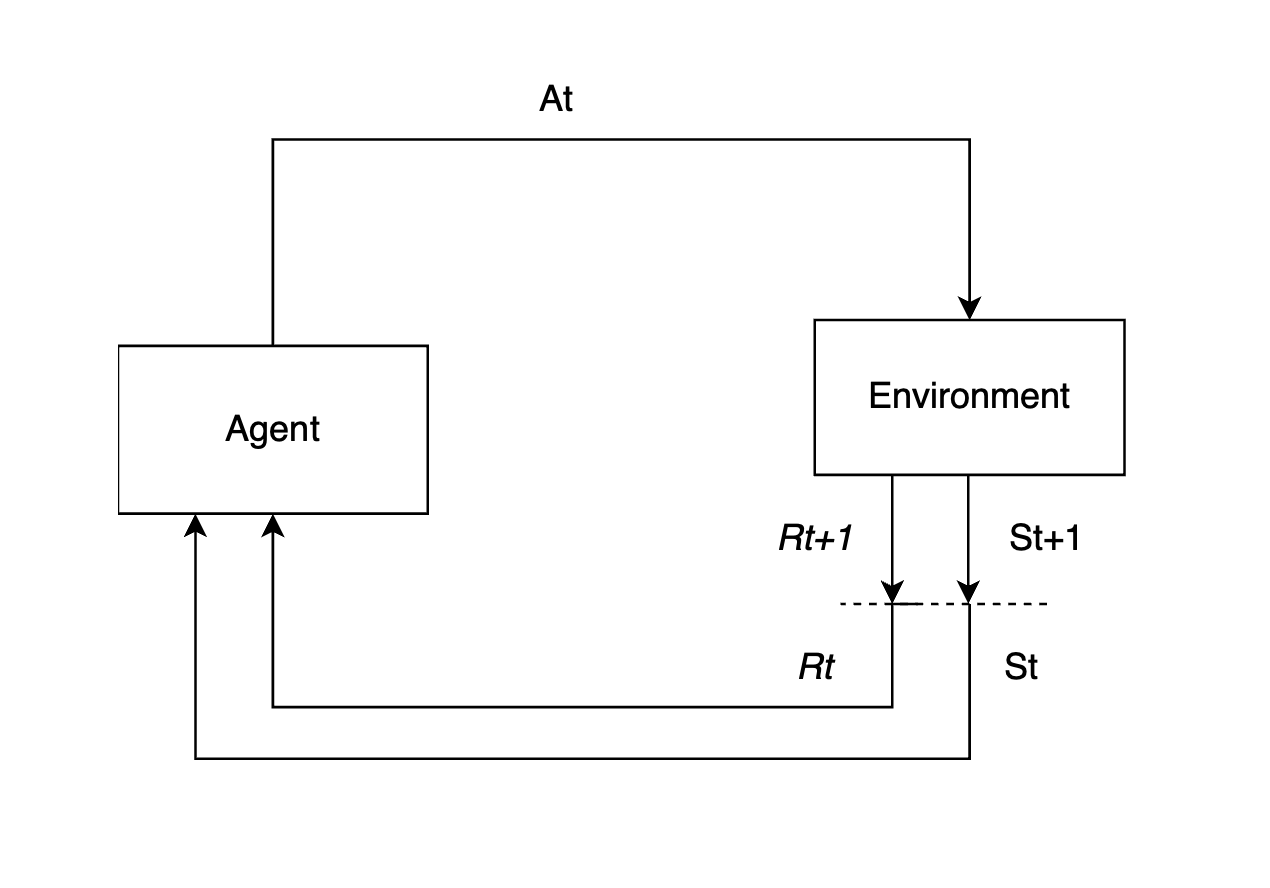}
    \caption{The RL interaction loop: At each time step $t$, the agent observes state $S_t$ and reward $R_t$, selects action $A_t$, and the environment responds with new state $S_{t+1}$ and reward $R_{t+1}$.}
    \label{fig:rl_loop}
\end{figure}

An RL problem consists of an \textit{agent} (the decision-making component) and an \textit{environment} (the system being controlled). At each time step $t$, the agent observes state $S_t$, selects action $A_t$, and receives reward $R_{t+1}$ along with the next state $S_{t+1}$. This cycle continues, typically in episodes terminating upon goal achievement or failure. Figure~\ref{fig:rl_loop} illustrates this interaction loop.

{RL problems are formalized as Markov decision processes (MDPs), defined here by the tuple $(\mathcal{S},\mathcal{A},p,r,\gamma)$, where $\mathcal{S}$ is the state space, $\mathcal{A}$ is the action space, $p(s',r\mid s,a)$ gives the joint distribution of the next state and reward, $r(s,a)$ denotes the expected immediate reward when this expectation is used explicitly, and $\gamma\in[0,1)$ is the discount factor. The Markov property requires
\[
P(S_{t+1},R_{t+1}\mid S_0,A_0,\ldots,S_t,A_t)
=
P(S_{t+1},R_{t+1}\mid S_t,A_t),
\]
meaning that, conditioned on the current state and action, the next state and reward are independent of the earlier history.}

The agent's goal is to learn a policy $\pi(a \mid s)$ maximizing expected cumulative discounted return:
\begin{equation}
G_t = \sum_{k=0}^{\infty} \gamma^k R_{t+k+1}.
\end{equation}

For beamline control, we employ the Deep Deterministic Policy Gradient (DDPG) algorithm~\cite{lillicrap2015}, which extends the deterministic policy gradient framework of Silver~\textit{et al.}~\cite{silver:hal-00938992} to deep neural networks. DDPG is an actor--critic method combining policy gradients for continuous actions with value estimation for stability. DDPG is model-free in the RL sense as it does not learn an internal dynamics model but instead relies on direct interaction with the simulator and uses experience replay and target networks to improve training stability and sample efficiency.

\subsection{Elegant Simulation Program}
\label{sec:elegant}

Simulation environments are preferred over real-world setups for RL training due to efficiency, safety, and cost-effectiveness - all the more so for particle accelerators, where misconfigured magnets can damage equipment. They also provide greater flexibility, allowing beamline layouts and element types to be modified and tested easily without changes to physical hardware. Among available tools such as MAD-X~\cite{grote2003madx} and ASTRA~\cite{floettmann2017astra}, we select Elegant~\cite{borland2000} for its flexible particle tracking and support for a wide range of accelerator elements.

\textbf{Elegant} (ELEctron Generation ANd Tracking) is a simulation code widely used for modeling charged particle beams in circular and linear accelerators. At its core, Elegant tracks particle motion in six-dimensional phase space:
\begin{equation}
\mathbf{x} = (x, x', y, y', s, \delta)^\top,
\end{equation}
where:
\begin{itemize}
    \item $x, y$ are transverse displacements (m) from the reference trajectory
    \item $x' = dx/ds \approx p_x/p_z$, $y' = dy/ds \approx p_y/p_z$ are dimensionless transverse slopes
    \item $s$ is the path length (m) along the reference trajectory
    \item $\delta = (p - p_0)/p_0$ is the fractional momentum deviation from reference momentum $p_0$
\end{itemize}

This 6D representation is complete for single-particle dynamics in the paraxial approximation: given $\mathbf{x}(s_0)$ and beamline element parameters, $\mathbf{x}(s_1)$ is deterministically computable. For linear optics, the transformation is:
\begin{equation}
\mathbf{x}(s_1) = {\mathbf{M}(s_0 \to s_1)} \mathbf{x}(s_0),
\end{equation}
where {$M$} is the $6 \times 6$ transfer matrix. {Elegant can model nonlinear dynamics using higher-order maps, canonical-kick elements, or numerically integrated element models~\cite{borland2000}.}

Elegant's inputs are structured in two files: the \texttt{.lte} file (lattice file) describes the beamline layout including magnets, drifts, apertures, RF cavities, and diagnostic points; the \texttt{.ele} file specifies beam properties (initial distributions, energy, particle count) and simulation instructions. Outputs are in SDDS format, a flexible binary structure with self-describing headers, processed using the SDDS Toolkit for analysis and visualization.

\section{Methodology}
\label{sec:methodology}

We formulate beamline optimization as a Markov Decision Process (MDP) solved using reinforcement learning. We build on the RL structure depicted in Figure~\ref{fig:rl_loop}, where an agent selects actions and an environment responds by executing those actions, running a beamline simulation, and providing feedback as a new {observation}, reward, and episode completion signal.

{ Beamline tuning has a feedback character: operators observe diagnostics, adjust magnet settings, and observe the response. RLABC formulates this process as a sequential decision problem. At step $t$, the agent observes beam information from the watch point immediately upstream of the current tunable element, applies an action to that element, and the environment reruns Elegant with the accumulated magnet settings to obtain the next observation and reward.}

{
\subsection{Markov Property Justification}
\label{sec:markov}

At decision step $t$, let
\begin{equation}
\mathbf{x}_{i,t}=(x_{i,t},x'_{i,t},y_{i,t},y'_{i,t},s_{i,t},\delta_{i,t})^\top\in\mathbb{R}^6
\end{equation}
denote the six-dimensional phase-space coordinate of surviving particle $i$ at the watch point immediately upstream of the $t$-th tunable element. The particle-distribution component of the simulator state is
\begin{equation}
\label{eq:Xt}
X_t=\{\mathbf{x}_{i,t}\}_{i=1}^{N_t},
\end{equation}
where $N_t$ is the number of particles that have survived up to decision step $t$.

The action $a_t\in\mathbb{R}^4$ sets the parameters of the current tunable element according to its type (Table~\ref{tab:actions}). Previously tuned upstream elements retain their assigned values; their effect on later decisions is represented through the particle distribution $X_t$ observed at the current watch point. Beam propagation from the current watch point to the next one is then described by the step-specific Elegant transport map
\begin{equation}
\label{eq:Ft}
X_{t+1}=F_t(X_t,a_t;\Theta_t),
\end{equation}
where $\Theta_t$ contains the fixed parameters of the intervening lattice segment, including drifts, apertures, RF settings, and non-tuned element parameters. During this transport, particles that hit apertures or leave the accepted region are removed from the surviving ensemble; hence the survivor count can only stay the same or decrease, $N_{t+1}\leq N_t$.

The reward is computed from the particle survival across this segment,
\begin{equation}
r_{t+1}=R_t(N_{\mathrm{ref}},N_t,N_{t+1};M,N_{\mathrm{loss}}),
\end{equation}
with the explicit form given in Eq.~\ref{eq:reward}. An episode terminates when the number of surviving particles falls below the loss threshold, $N_{t+1}\leq N_{\mathrm{loss}}$, or when the final watch point is reached. Here $t$ denotes the decision index, or equivalently the current beamline step, not physical clock time. Because different decision steps correspond to different lattice segments, the transition rule is step-dependent. Thus the simulator defines a finite-horizon Markov process with transition kernel $P_t$ at step $t$: conditioned on the current particle distribution $X_t$, action $a_t$, and known step index $t$, the next distribution and reward are independent of earlier history:
\begin{equation}
\label{eq:markov}
P_t(X_{t+1},r_{t+1}\mid X_0,a_0,\ldots,X_t,a_t)
=
P_t(X_{t+1},r_{t+1}\mid X_t,a_t).
\end{equation}

\textbf{MDP summary.} The full simulator formulation is a finite-horizon Markov process with decision index $t$, simulator state $S_t\equiv X_t$, action $A_t\equiv a_t\in\mathcal{A}\subset\mathbb{R}^4$ with element-dependent masking, {effective action $\widetilde{A}_t=m_t\odot A_t$ for a fixed four-dimensional element-type mask $m_t$,} transition kernel $P_t$ induced by Elegant tracking through $F_t$, reward $R_t$ defined in Eq.~\ref{eq:reward}, and discount factor $\gamma=0.99$.

}

\begin{figure}[!ht]
\centering
\includegraphics[width=0.9\textwidth]{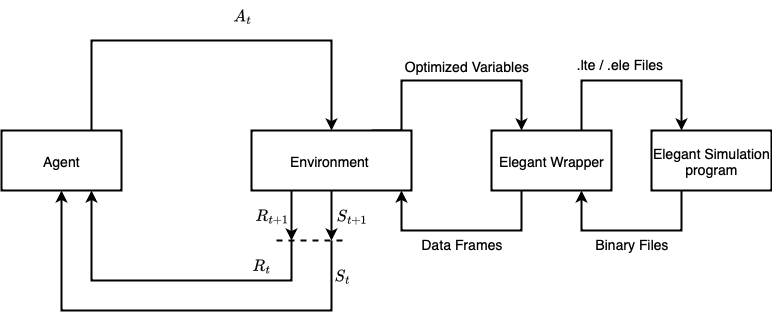}
\caption{System architecture of RLABC: The RL agent interacts with the environment, which communicates with the Elegant Wrapper and simulation engine.}
\label{fig:system_layout}
\end{figure}

For simulations, we use Elegant, interfaced through a specialized Elegant Wrapper that streamlines Python integration. The full RLABC architecture is shown in Figure~\ref{fig:system_layout}.

\subsection{Elegant Wrapper}
\label{sec:wrapper}

The Elegant Wrapper serves as a bridge between Python code and the Elegant simulation tool, processing the \texttt{.lte} (lattice) and \texttt{.ele} (command) files with the following capabilities:

\begin{itemize}
\item \textbf{Beamline parsing:} Reads input files and constructs a graph representation $G = (V, E)$ of the beamline, where vertices $V$ represent elements and edges $E$ represent sequential connections. This enables efficient queries such as identifying tunable elements between positions or finding downstream watch points.
\item \textbf{Beamline preprocessing:} Modifies the layout by inserting watch points immediately before each tunable element (quadrupoles, dipoles), creating a standardized structure: fixed sections (drifts, apertures) followed by a watch point, then a tunable element. {This placement enables observation of beam conditions immediately before each simulated decision point and supports the finite-horizon MDP formulation described in the previous subsection.} Figure~\ref{fig:beamline_preprocessing} illustrates this transformation.

\begin{figure}[!ht]
\centering
\includegraphics[width=0.9\textwidth]{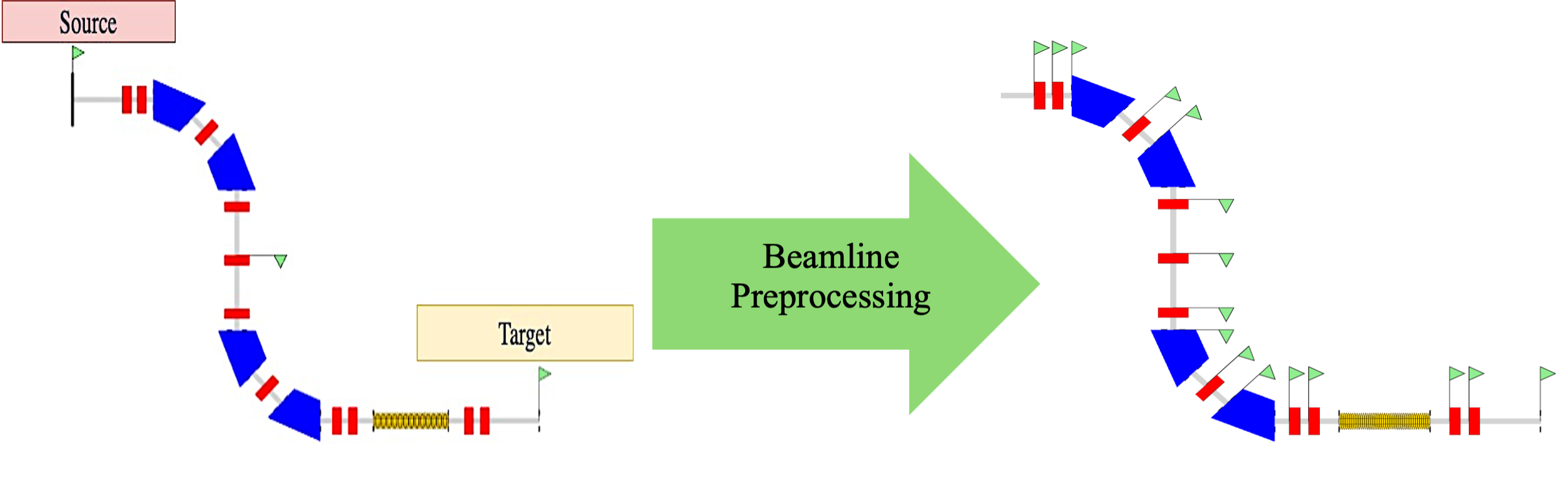}
\caption{Beamline architecture before (left) and after (right) preprocessing. Blue blocks: dipole magnets; red blocks: quadrupoles; green flags: watch points; yellow block: RF cavity. After preprocessing, a watch point appears before every controllable element.}
\label{fig:beamline_preprocessing}
\end{figure}

\item \textbf{Simulation control:} Allows customization of parameters (particle count, lattice values) and executes Elegant simulations.

\item \textbf{Data handling:} Converts binary SDDS output files to Python formats (pandas DataFrames) for RL system integration{~\cite{mckinney2010pandas}}.

\item \textbf{Visualization:} Provides plotting functions for accelerator physics quantities including beta functions ($\beta$), dispersion ($D$), emittance ($\epsilon$), and Twiss parameters{~\cite{courant1958,wiedemann2015}}.
\end{itemize}

\subsection{Environment}
\label{sec:Environment}

{Using the Elegant Wrapper as foundation, the RL environment manages simulation-agent interaction following the Gymnasium interface~\cite{2025gymnasiums}.} Algorithm~\ref{alg:environment_episodes} describes the episode procedure.

\begin{algorithm}[H]
\caption{RLABC Environment Episode}\label{alg:environment_episodes}
\begin{algorithmic}[1]
\Procedure{RunEpisode}{{$G$}, {$N_0,N_{\mathrm{loss}}=3$}}
    \State {$M \gets \text{number\_of\_controllable\_elements}(G)$}
    \State $w_0 \gets \text{first\_watch\_point}(G)$
    \State $\text{run\_elegant}()$ \Comment{Run Elegant simulation}
    \State {$o_0 \gets \text{extract\_observation}(w_0)$} \Comment{{57-element vector}}
    \State $t \gets 0$
    
    \While{$t < M$} \Comment{$M$ = number of tunable elements}
        \State $e_t \gets \text{next\_tunable\_element}(w_t, G)$
        \State $w_{t+1} \gets \text{next\_watch\_point}(e_t, G)$
        
        \State $a_t \gets \text{agent.select\_action}({o_t})$ \Comment{RL agent}
        \State {$m_t \gets \text{action\_mask}(e_t)$} \Comment{{Fixed 4D type mask}}
        \State {$\widetilde{a}_t \gets m_t \odot a_t$} \Comment{{Effective policy-space action}}
        \State $\text{apply\_action}(e_t, {\widetilde{a}_t})$ \Comment{Update .lte file}
        
        \State $\text{run\_elegant}()$ \Comment{Run simulation}
        \State $N_{t+1} \gets \text{count\_particles}(w_{t+1})$
        
        \State {$o_{t+1} \gets \text{extract\_observation}(w_{t+1})$}
        \State $r_{t+1} \gets \text{compute\_reward}(N_t, N_{t+1}, N_0, t, M)$
        \State $\text{done} \gets (N_{t+1} \leq {N_{\mathrm{loss}}}) \lor (t{+}1=M)$ \Comment{{$N_{\mathrm{loss}}=3$ particles}}
        
        \State {$m_{t+1} \gets \text{next\_action\_mask}(w_{t+1},G)$}
        \State {$\text{agent.store}(o_t, \widetilde{a}_t, r_{t+1}, o_{t+1}, \text{done},m_t,m_{t+1})$}
        \If{\text{done}} \textbf{break}
        \EndIf
        \State {$o_t \gets o_{t+1}$}, $t \gets t + 1$
    \EndWhile
\EndProcedure
\end{algorithmic}
\end{algorithm}

\subsubsection{{Observation Representation}}

{The policy receives a fixed 57-dimensional observation $O_t=\phi_t(X_t)$ extracted from watch-point data. Because this feature map compresses the full particle distribution $X_t$, the policy operates under partial observability.} {The fixed dimension ensures compatibility with standard neural network architectures regardless of the number of surviving particles at each step.}

{The observation combines six groups of features. Sixteen entries contain the median, interquartile range, and 10th and 90th percentiles of $x$, $y$, $x'$, and $y'$, providing robust summaries of transverse location and spread. A normalized $5\times5$ histogram contributes 25 entries describing the transverse $x$--$y$ distribution. Two scalar entries give the surviving-particle fraction $N_t/N_{\mathrm{ref}}$ and the fraction of particles inside the upstream aperture ellipse. Ten entries contain the upper triangle of the correlation matrix of the standardized coordinates $(x,y,x',y')$, representing linear coupling among the transverse coordinates. The final four entries are the upstream and downstream aperture semi-axes, $A_x^{\mathrm{before}}$, $A_y^{\mathrm{before}}$, $A_x^{\mathrm{after}}$, and $A_y^{\mathrm{after}}$. Thus, the observation combines distribution shape, particle survival, transverse coupling, and local geometric constraints in a fixed-size vector.}

\subsubsection{Action Space}
\label{sec:action_space}

At each step the agent outputs a 4-dimensional continuous vector $a_t \in \mathbb{R}^4$. {The interpretation of its components depends on the element type at step $t$: for quadrupoles, the first three components are used ($K_1$, HKICK, VKICK) and the fourth is ignored; for dipoles, only the last component (FSE) is active and the first three are masked.} This uniform action dimensionality simplifies the neural network architecture by avoiding variable-size outputs, while the masking ensures that inactive dimensions do not affect the simulation. {The mask is applied both by the environment and inside the actor--critic update, as detailed below.} Table~\ref{tab:actions} summarizes the action space parameters.

\begin{table}[!ht]
\caption{Action space parameters. For quadrupole steps, the first three action dimensions are used; for dipole steps, only the last dimension (FSE) is applied.}
\label{tab:actions}
\centering
\begin{tabular}{llll}
\toprule
\textbf{Element} & \textbf{Parameter} & \textbf{Range} & \textbf{Description} \\
\midrule
\multirow{3}{*}{Quadrupole} & $K_1$ & {$[-20, 20]$ m$^{-2}$} & Focusing strength \\
 & HKICK & $[-0.005, 0.005]$ rad & Horizontal kick \\
 & VKICK & $[-0.005, 0.005]$ rad & Vertical kick \\
\midrule
Dipole & FSE & $[-0.005, 0.005]$ & Fractional strength error \\
\bottomrule
\end{tabular}
\end{table}

{\textbf{Masking in the actor--critic pipeline.} At decision step $t$, the current element type defines a binary mask $m_t\in\{0,1\}^4$ over the fixed policy action $a_t=[K_1,\mathrm{HKICK},\mathrm{VKICK},\mathrm{FSE}]$. For a quadrupole, $m_t=[1,1,1,0]$; for a dipole, $m_t=[0,0,0,1]$. The environment applies this active-component rule before converting the policy output to the physical settings of the current element, and the actor-critic pipeline applies the corresponding four-dimensional mask to the same action components. Consequently, parameters that cannot affect the simulated transition are also excluded from value estimation and policy optimization. The effective action $\widetilde{a}_t=m_t\odot a_t$ is stored in replay together with the current mask $m_t$ and the next-element mask $m_{t+1}$. The critic $Q(o,a)$ estimates the expected discounted return after action $a$ is taken from observation $o$ and the policy is followed thereafter~\cite{sutton2018,lillicrap2015}. The online critic is therefore fitted using $Q(o_t,\widetilde{a}_t)$. Here, $\mu$ denotes the online actor and $\mu'$ denotes its target-network copy. For the Bellman target, the target-policy action is masked as $\widetilde{a}_{t+1}'=m_{t+1}\odot\mu'(o_{t+1})$ before evaluation by the target critic. Likewise, the actor objective evaluates $Q(o_t,m_t\odot\mu(o_t))$, so inactive action components receive no direct policy gradient. Because these masks depend on element type rather than beamline length, the same actor and critic architecture can be used for supported beamlines with different numbers or orderings of tunable elements.}

\subsubsection{Reward Function}
\label{sec:reward_function}

The reward encourages high transmission while penalizing early beam loss. With $t$ as the {completed element step}, {$N_{\mathrm{ref}}$ as the particle count at the initial watch point}, $N_t$ current particles, $N_{t-1}$ previous particles, $M$ total elements, and base transmission {$B_t=N_t/N_{\mathrm{ref}}$}:

\begin{equation}
\label{eq:reward}
R_t =
\begin{cases}
B_t - \dfrac{\sqrt{M^2 - t^2}}{M}, & \text{if } N_t \leq {N_{\mathrm{loss}}}, \\[10pt]
B_t \cdot \dfrac{N_t}{N_{t-1}}, & \text{if } N_t > {N_{\mathrm{loss}}}.
\end{cases}
\end{equation}

{where $N_{\mathrm{loss}}=3$ is the early-termination threshold used in the reported code. In the code, the correlation matrix used in the observation cannot be calculated with fewer than three particles. We are assuming that if three or fewer particles remain, almost the whole beam has been lost, so the episode has failed.}

This design provides:
\begin{itemize}
\item \textbf{Global transmission feedback} ($B_t$ term): Links reward to surviving fraction
\item \textbf{Local retention bonus} ($N_t/N_{t-1}$ term): Encourages minimizing losses at each step
\item \textbf{Early loss penalty}: Larger penalties for losing particles early in the beamline (when $t \ll M$)
\end{itemize}

{Episodes terminate when $N_t\leq N_{\mathrm{loss}}$ or when the beam reaches the final watch point at the end of the modeled beamline.}

\subsection{Agent}
\label{sec:agent}

{RLABC separates the environment from the learning algorithm. The environment follows the Gymnasium interface and can be used with Stable-Baselines3~\cite{raffin2021}. In the current RLABC interface for Stable-Baselines3, the mask is applied only on the environment side, before the action is sent to Elegant. The reported experiments use our DDPG implementation, which applies the mask both on the environment side and during the actor--critic update. Extending the Stable-Baselines3 interface to support this second mask will be explored in future work.}

Since the focus of this work is the environment construction methodology rather than algorithm selection, we adopt DDPG~\cite{lillicrap2015} as a representative algorithm {with four hidden layers (800, 600, 512, and 256 units) and layer normalization; the action is injected at the final hidden layer of the critic}. {The actor and critic learning rates are $10^{-4}$ and $10^{-3}$, respectively, and exploration uses Gaussian noise.} {The reported DDPG hyperparameters are summarized in Table~\ref{tab:hyperparams}, and the per-experiment training lengths and exploration schedules are given in Section~\ref{sec:reproducibility}.} Systematic comparison across algorithms and hyperparameter configurations is left to future work.

\subsection{Stage Learning}
\label{sec:stage_learning}

To create a generalized and scalable training framework for complex beamline configurations, we introduce \textit{stage learning} strategies that simplify the optimization problem and improve training efficiency.

Stage learning decomposes a complex task into progressively more challenging subtasks. An agent is first trained on a simplified problem, then the learned policy parameters and replay buffer serve as warm starts for subsequent stages. Two complementary forms are implemented:

\textbf{Beamline segmentation:} Given a beamline with multiple elements, stages incrementally increase the number of active elements. For example, with ten elements, an initial stage may involve only the first three. Once converged, trained weights and replay buffer initialize training for the next stage with additional elements. This continues until the full beamline is optimized.

\textbf{Action space progression:} Different parameters operate on significantly different scales. Quadrupole strengths ($K_1$) lie within {$[-20,20]$}, while correction parameters (FSE, VKICK, HKICK) are constrained to {$[-5 \times 10^{-3},5 \times 10^{-3}]$}. The framework allows initially optimizing only a subset (e.g., $K_1$ values) before progressively introducing additional control variables.

{Both strategies are optional and can be used independently or jointly; training from scratch remains supported. In the completed three-run comparison reported in Section~\ref{sec:multiseed_stage}, a short K-only warm start improves early adaptation after all variables are activated, while the selected-policy endpoint is similar to training from scratch.}

\section{Testing the algorithm with a realistic beamline model}
\label{sec:results}

\begin{figure}[!ht]
    \centering
    \includegraphics[width=0.9\textwidth]{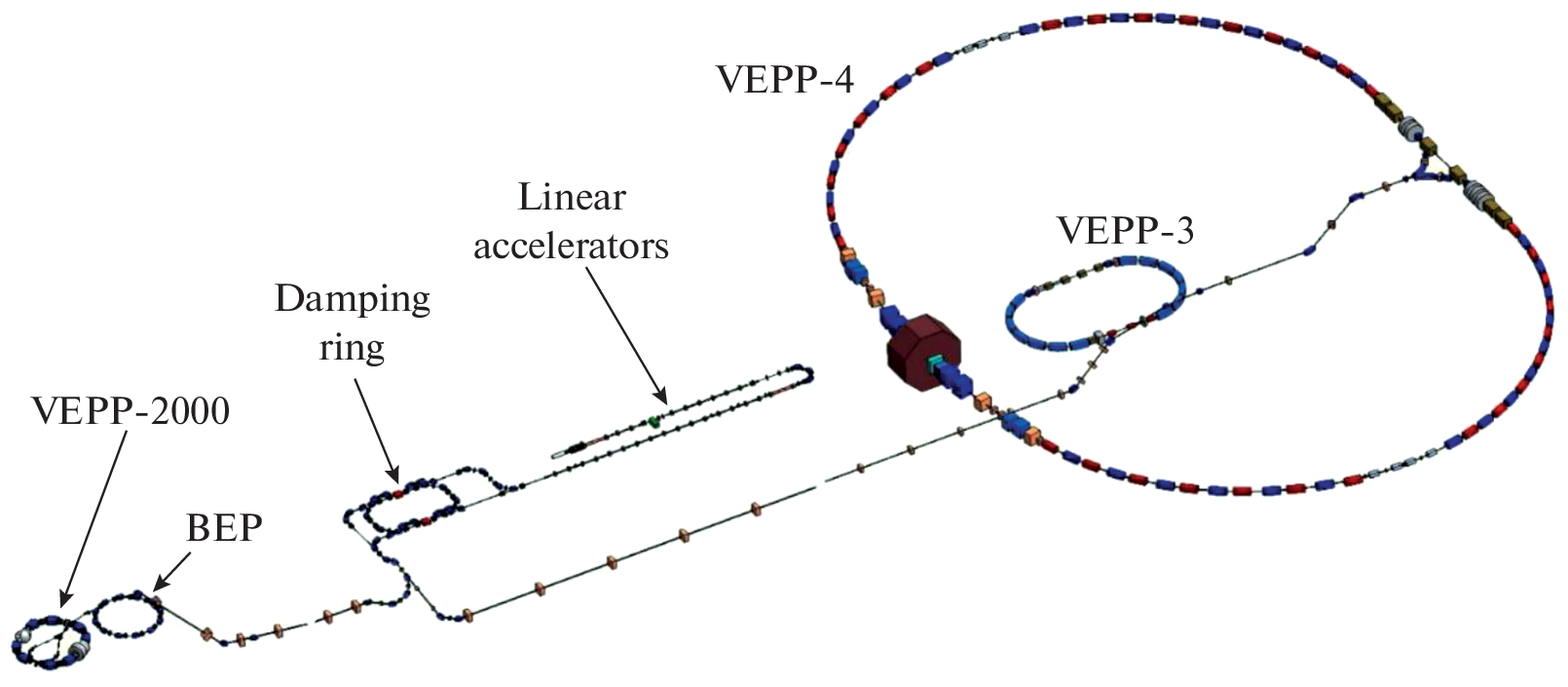}
    \caption{The VEPP-5 injection complex~\cite{Emanov2023-wx}.}
    \label{fig:VEPP-5InjectionComplex}
\end{figure}

\begin{figure}[!ht]
    \centering
    \includegraphics[width=0.9\textwidth]{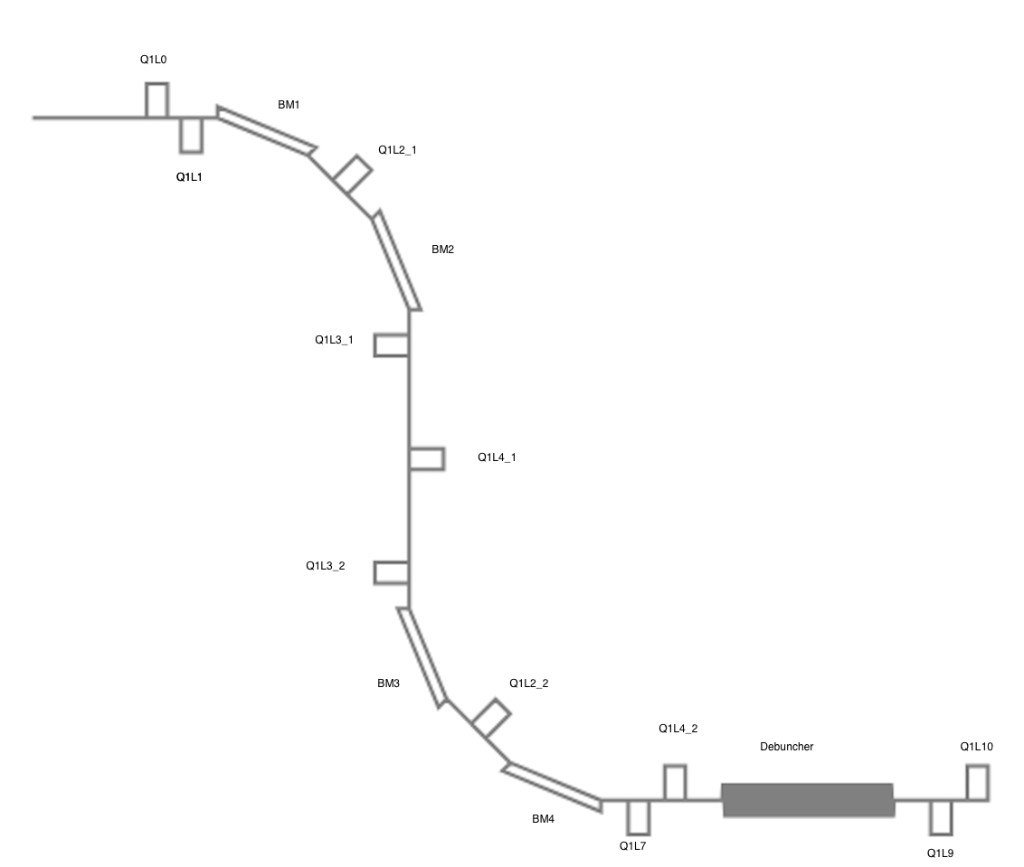}
    \caption{The test beamline layout: a simplified segment of the VEPP-5 positron transport channel, comprising 11 quadrupoles (Q1L0--Q1L10), 4 dipoles (BM1--BM4), and an RF debuncher cavity (37 tunable parameters).}
    \label{fig:optimized_beamline}
\end{figure}

We evaluate RLABC on a test beamline derived from the VEPP-5 injection complex at BINP~\cite{Emanov2023-wx}. Figure~\ref{fig:VEPP-5InjectionComplex} shows the full facility, which includes two linacs (electron and positron), injection and extraction channels, and a damping ring. The segment we use as our test beamline is a simplified model of the positron transport channel, treating only the horizontal plane (vertical dispersion is omitted). Figure~\ref{fig:optimized_beamline} shows the test beamline layout: 11 quadrupole magnets, 4 dipoles, and an RF cavity, yielding 37 tunable variables.

\subsection{Setup of the tests}
\label{sec:test_setup}

{The test-beamline experiments are run on a MacBook Pro with an Apple M3 Pro chip and 18 GB RAM.} The two-dipole beamline variant (Section~\ref{sec:two_dipole}) was trained on a workstation equipped with an Intel Core i7-8700K CPU (3.70 GHz, 6 cores/12 threads), 32 GB RAM, and an NVIDIA GeForce GTX 1060 6GB GPU. {The observed single-simulation wall times at $N_0=10^3$, $10^4$, and $10^5$ are, respectively, 0.042, 0.261, and 2.959~s on the MacBook, and 1.155, 1.395, and 2.515~s on the Linux workstation. A complete episode with $M$ controllable elements requires $M+1$ Elegant simulations: one for each of the $M$ sequential element updates and one additional simulation at the final watch point at the end of the optimized beamline. The test beamline has $M=15$, so a full-beamline episode uses 16 simulations, whereas an episode that fails because all particles are lost before reaching the end uses fewer. Direct end-to-end measurements on the MacBook, including the environment, Elegant, and SDDS processing, give $2.36\pm0.01$, $6.97\pm0.18$, and $39.61\pm6.15$~s for $N_0=10^3$, $10^4$, and $10^5$, respectively (mean $\pm$ sample SD over three complete episodes).}

At the start of each episode, the environment's reset function initializes all magnet parameters (quadrupole strengths $K_1$, corrector kicks HKICK/\allowbreak VKICK, and dipole FSE values) to zero, ensuring that the agent discovers configurations from a neutral starting point without any prior knowledge of design optics. {Training and periodic noise-free evaluation episodes use $N_0=10^3$ particles. Every selected configuration in the comparison below is subsequently re-evaluated with $N_0=10^5$.}

\subsection{Training Dynamics and Optimization Objective}

An important distinction between beamline \textit{optimization} and real-time \textit{control} is the relevant performance metric. In real-time control applications, consistent average performance matters because the policy must perform reliably on every deployment. In contrast, beamline optimization aims to \textit{discover} high-transmission magnet configurations: once a sufficiently good configuration is found, those specific parameter values (quadrupole strengths $K_1$, corrector kicks HKICK/VKICK, dipole FSE) can be extracted for use. The RL agent serves as an optimization tool, not a deployed controller.

\begin{figure}[!ht]
\centering
\subfloat[Test beamline (37 parameters)]{\includegraphics[width=0.48\textwidth]{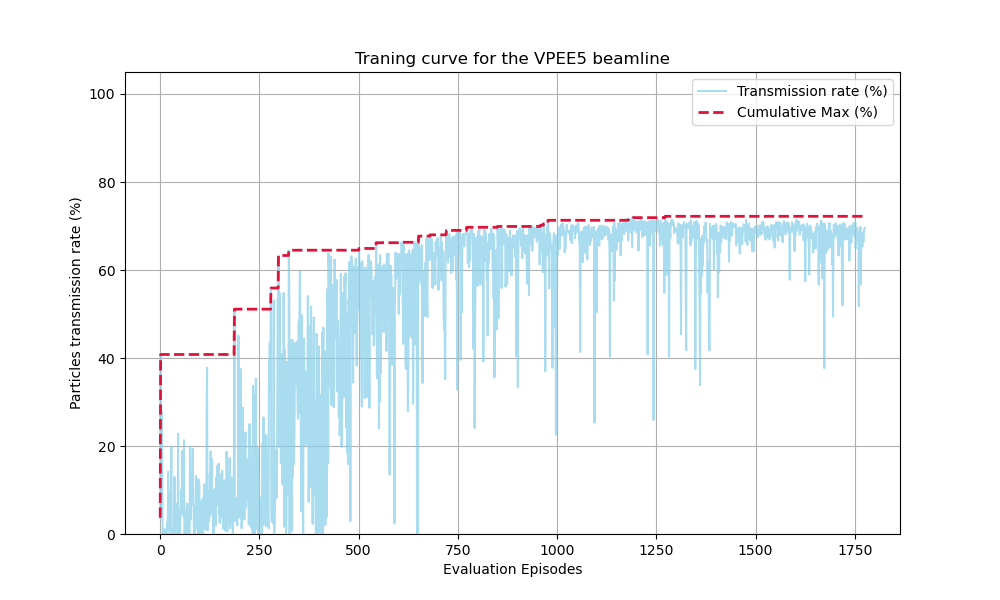}\label{fig:training_curve_vepp5}}
\hfill
\subfloat[Two-dipole variant (35 parameters)]{\includegraphics[width=0.48\textwidth]{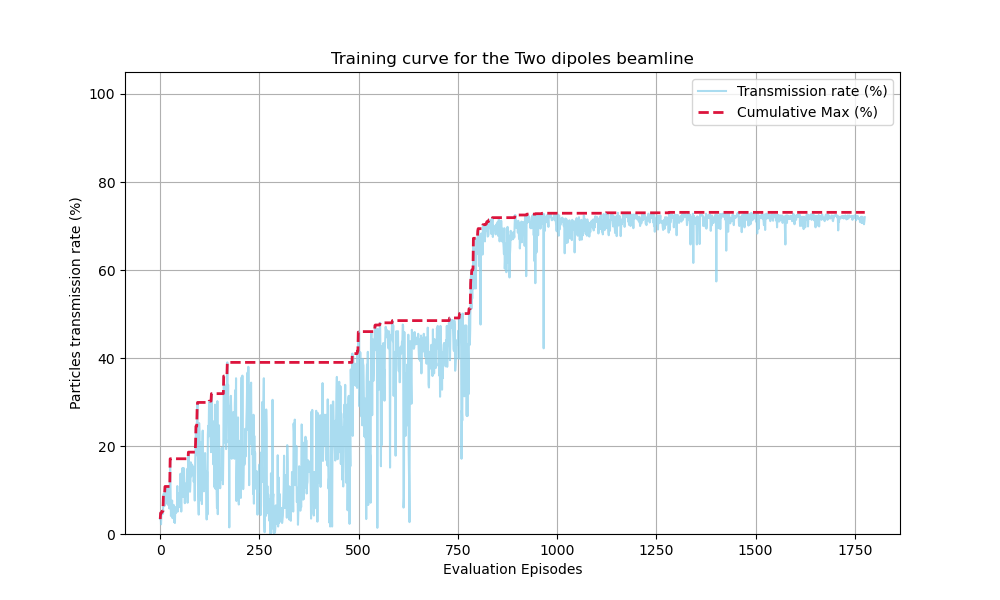}\label{fig:training_curve_2d}}
\caption{{Training curves showing DDPG agent performance during evaluation episodes. Light blue traces show individual episode transmission rates; the dashed red line tracks the cumulative maximum. (a) Test beamline. (b) Two-dipole variant (Section~\ref{sec:two_dipole}).}}
\label{fig:training_curves}
\end{figure}

Figure~\ref{fig:training_curves} shows the learning progress during training. The agent explores the 37-dimensional parameter space, with individual episodes (light blue) showing high variance as different configurations are tested. The cumulative maximum (dashed red line) tracks the best transmission achieved up to each episode. What matters for deployment is the \textbf{best configuration} found during training, i.e. the parameter set extracted for use. {The same selection principle is used for the classical optimizers: each independent run contributes its best observed configuration rather than its average over all objective evaluations.}

For comparison, Figure~\ref{fig:training_curve_2d} shows the training curve for the two-dipole beamline variant discussed in Section~\ref{sec:two_dipole}. Both configurations exhibit qualitatively similar learning dynamics, with the agent progressively discovering higher-transmission configurations over the course of training.

\subsection{{Optimization Results}}
\label{sec:multiseed_stage}

{To evaluate reproducibility, every experiment summarized by a mean and sample standard deviation in Table~\ref{tab:multiseed_validation} is repeated with three independent numerical seeds. During training, each DDPG policy is evaluated periodically without exploration noise, providing intermediate measurements of policy performance. Each classical-optimizer run likewise contributed its best observed configuration. Every selected configuration is re-evaluated with $N_0=10^5$ particles; the optimizer-specific selection particle counts and search budgets are stated below.}

{The classical comparators are SciPy's population-based differential-evolution implementation~\cite{Storn1997-db,2020SciPy-NMeth} and the \texttt{bayesian-optimization} package~\cite{Nogueira2014BayesianOptimization}, using its Gaussian-process optimizer with an upper-confidence-bound acquisition~\cite{pelikan2005bayesian}. They use the same Elegant wrapper and physical parameter bounds as DDPG.}

\begin{table}[!ht]
\caption{{Selected-configuration validation across three independently initialized runs per experiment. The third column gives the mean transmission in percent $\pm$ the sample SD in percentage points; the final column gives the best independently re-evaluated transmission among the three runs.}}
\label{tab:multiseed_validation}
\centering
\small
{%
\resizebox{\textwidth}{!}{%
\begin{tabular}{@{}llcc@{}}
\toprule
\textbf{Method and experiment} & \textbf{Active variables} & \textbf{Mean $\pm$ SD (\%)} & \textbf{Best (\%)} \\
\midrule
DDPG, K-only & 11 $K_1$ & $69.8\pm0.7$ & 70.6 \\
Differential evolution (K-only) & 11 $K_1$ & $62.9\pm7.7$ & 70.7 \\
Bayesian optimization (UCB; K-only) & 11 $K_1$ & $4.3\pm1.2$ & 5.5 \\
\addlinespace
DDPG, short K-only warm start & All 37 & $66.3\pm3.7$ & 68.7 \\
DDPG, from scratch & All 37 & $66.2\pm3.4$ & 68.8 \\
Differential evolution (all variables) & All 37 & $1.8\pm0.6$ & 2.4 \\
Bayesian optimization (UCB; all variables) & All 37 & $0.2\pm0.1$ & 0.2 \\
\bottomrule
\end{tabular}%
}%
}
\end{table}

{Table~\ref{tab:multiseed_validation} summarizes both the quality and the run-to-run consistency of the selected configurations. In the K-only search space, DDPG reaches a mean transmission of 69.8\%, with a sample SD of 0.7 percentage points, compared with a mean of 62.9\% and a sample SD of 7.7 percentage points for differential evolution. The best configurations found by the two methods are comparable, reaching 70.6\% and 70.7\%, respectively. However, the higher DDPG mean and substantially smaller standard deviation show that the completed DDPG runs consistently produced high-transmission configurations across the independent initializations. The independently initialized Bayesian-optimization experiments obtained a mean of 4.3\%, a sample SD of 1.2 percentage points, and a best transmission of 5.5\%.}

{When the search is expanded to all 37 variables, DDPG produces similar selected-configuration results with and without the short K-only warm start: the mean transmissions are 66.3\% and 66.2\%, the sample SDs are 3.7 and 3.4 percentage points, and the best transmissions are 68.7\% and 68.8\%, respectively. The lower results obtained by differential evolution and independently initialized Bayesian optimization show that the expanded, coupled search space is more difficult to optimize under the settings and finite budgets used here. The initialization of a Bayesian-optimization run matters a great deal in problems of this size. Bayesian optimization seeds its surrogate model with an \textit{initial design} of evaluated points, and the accelerator-physics literature reports that supplying physics-informed prior information, or initial points already known to lie near a good solution, markedly improves convergence in high-dimensional tuning tasks~\cite{rousseletal2024,hanuka2021}. We therefore repeat the all-variable Bayesian optimization with an informed initial design containing a known high-transmission configuration. That run ends at a validated transmission of 62.9\%, compared with 0.2\% for the independently initialized runs, so the initial design is what separates a usable result from an unusable one in this search space.}

\begin{figure}[!ht]
    \centering
    \includegraphics[width=0.95\textwidth]{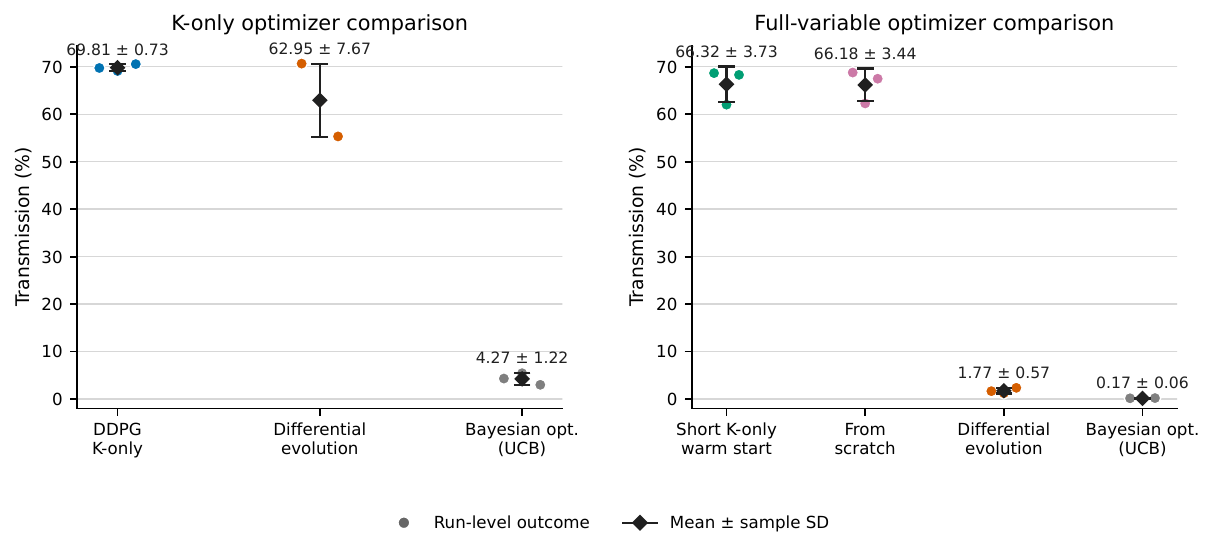}
    \caption{{Run-level selected-policy transmission in the high-statistics validation. Colored circles are the three independent outcomes for each experiment; black diamonds and error bars show the mean and $\pm$ one sample SD. Panel (a) compares methods in the common 11-variable K-only search space; panel (b) compares the DDPG and classical experiments in the 37-variable search space.}}
    \label{fig:n100000_stability}
\end{figure}

\begin{figure}[!ht]
    \centering
    \includegraphics[width=\textwidth]{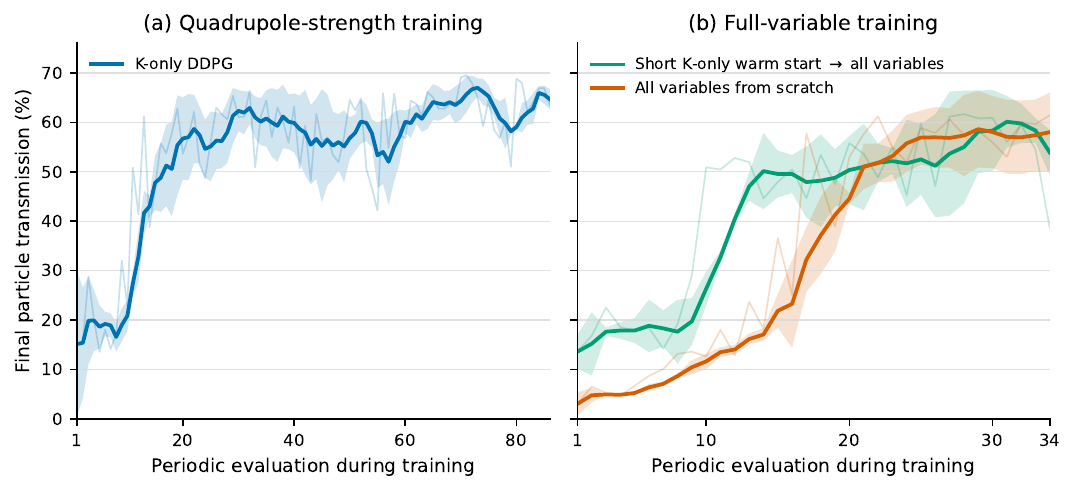}
    \caption{{DDPG evaluation performance during training. Faint curves show the mean across three independent runs, solid curves show a trailing moving average over five evaluations, and shaded bands indicate $\pm$ sample SD. Panel (a) presents K-only training; panel (b) compares warm-start and from-scratch full-variable training over their common first 34 evaluations.}}
    \label{fig:stage_multiseed}
\end{figure}

{From Table~\ref{tab:multiseed_validation} we see that for this beamline, the stage-learning warm start is not required to reach a similar selected-policy endpoint within the completed training and does not improve that endpoint in this comparison; its benefit is faster early adaptation after the full variable set is activated. For a longer or more complex beamline, earlier transport through a larger fraction of the lattice populates the replay buffer with more transitions from viable trajectories and reduces the time required to optimize the full beamline. Future work will test stage-learning approaches on beamlines with more controllable elements and larger coupled action spaces.}

\subsection{Parameter Convergence Analysis}
\label{sec:convergence}

\begin{figure}[!ht]
    \centering
    \includegraphics[width=\textwidth]{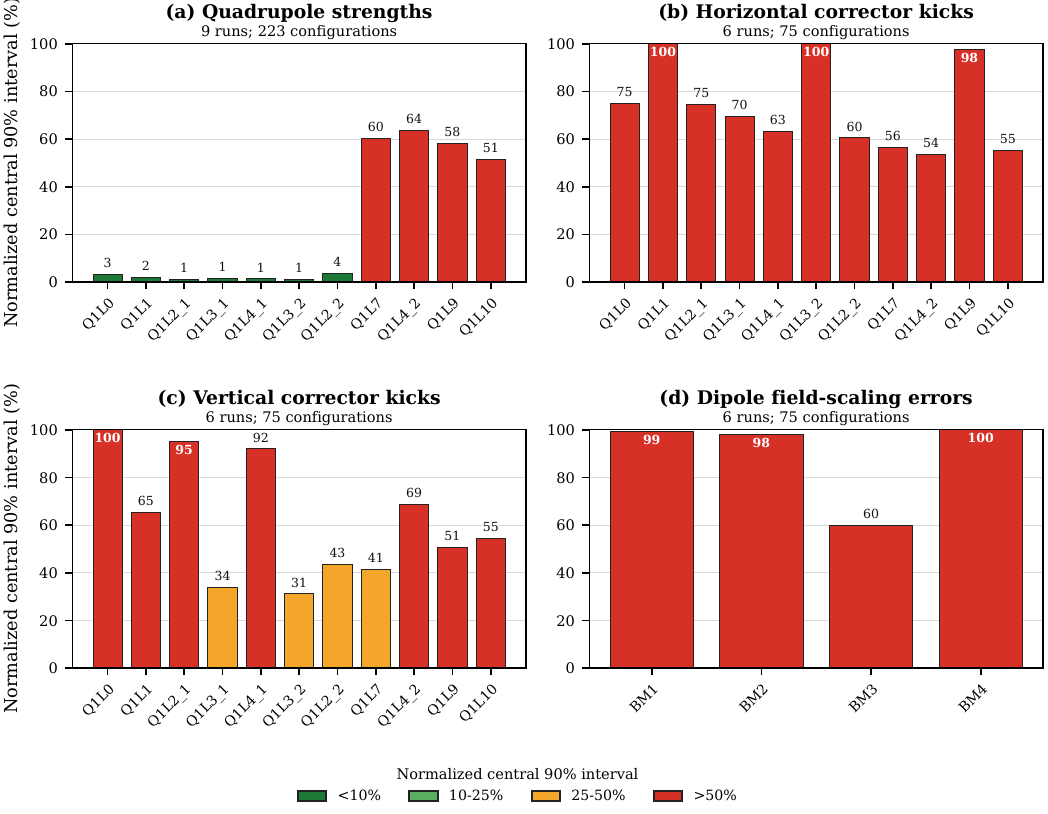}
    \caption{Convergence analysis of all 37 control parameters from high-performing evaluation episodes (transmission $>60\%$){, pooled over independent runs with equal weight per run. Bars show the fraction of each parameter's allowed range occupied by the middle 90\% of high-performing solutions}. Color indicates convergence strength: dark green ($<10\%$), green (10--25\%), orange (25--50\%), red ($>50\%$). Quadrupole strengths ($K_1$) show strong convergence, with {seven} of 11 parameters below 10\%. Corrector kicks show more variability, suggesting multiple viable steering solutions exist. {The $K_1$ panel uses 223 configurations from nine runs; the HKICK, VKICK, and FSE panels use 75 configurations from the six full-variable runs.}}
    \label{fig:param_all}
\end{figure}

Beyond aggregate transmission performance, we analyze whether the RL agent discovers consistent magnet configurations across independent runs or finds highly variable solutions each time. The test beamline optimization involves 37 control parameters: 11 quadrupole strengths ($K_1$), 11 horizontal corrector kicks (HKICK), 11 vertical corrector kicks (VKICK), and 4 dipole field scaling errors (FSE). Figure~\ref{fig:param_all} shows the spread of all parameters in high-performing \textit{evaluation} episodes (transmission $>60\%$){, pooled over all nine independent runs of the three DDPG experiments.}

{The convergence measure $F = (Q_{95}-Q_{5})/(u-l)\times100\%$, where $Q_{5}$ and $Q_{95}$ are the run-weighted 5th and 95th percentiles and $[l,u]$ is the allowed range,} quantifies the relative spread of each parameter across high-performing solutions. A low {$F$} indicates \textit{strong convergence}: the agent consistently discovers similar values for that parameter, implying the optimization landscape constrains it to a narrow high-performing region. A high {$F$} indicates \textit{weak convergence}: many different values appear in high-performing configurations, implying the optimization landscape is degenerate along that parameter - multiple settings yield comparable transmission. {We normalize by the allowed range rather than by the mean, as a coefficient of variation would, because the kicks and FSE values are centered on zero, where a mean-normalized measure becomes unstable; for the quadrupole strengths, whose means lie far from zero, the two measures agree to within a percentage point.}

\textbf{Findings by parameter type:}
\begin{itemize}
     \item \textbf{Quadrupole strengths ($K_1$)}: {Seven} of eleven quadrupoles converge tightly ({$F$: 1--4\%}), indicating that the focusing lattice is strongly constrained{: across all nine runs, the central 90\% interval of Q1L2\_1 spans only 13.6--14.1~m$^{-2}$ within bounds of $\pm20$~m$^{-2}$}. Only the {four} quadrupoles {downstream of the final dipole (Q1L7 and Q1L4\_2 before the debuncher cavity, Q1L9 and Q1L10 after it)} exhibit weak convergence ({$F$: 51--64\%}), revealing optimization degeneracy in the downstream section.
    \item \textbf{Corrector kicks (HKICK, VKICK)}: Most correctors show weak convergence ({$F>50\%$}), indicating the orbit correction problem is underdetermined - many steering solutions achieve comparable transmission. Notable exceptions include {the vertical kicks at the two S-bend quadrupoles Q1L3\_2 and Q1L3\_1 ($F$: 31\% and 34\%), in the same region where the quadrupole strengths are most tightly constrained}.
    \item \textbf{Dipole FSE}: {BM3FSE converges most strongly of the four ($F$: 60\%), while BM1FSE, BM2FSE, and BM4FSE occupy essentially their entire allowed ranges ($F$: 98--100\%)}, as their errors can be compensated by many downstream elements.
\end{itemize}

These results demonstrate that the RL agent, with no physics encoded in its reward function or network architecture, discovers solutions with a clear sensitivity structure: parameters are tightly constrained where precision is required and exhibit degeneracy where the problem permits multiple viable configurations. {Because the parameters vary jointly, this structure characterizes the discovered solutions rather than the response to varying one parameter in isolation.}

\subsubsection{Best Solution Analysis}
\label{sec:best_solution}

\begin{figure}[!htbp]
    \centering
    \includegraphics[width=0.85\textwidth]{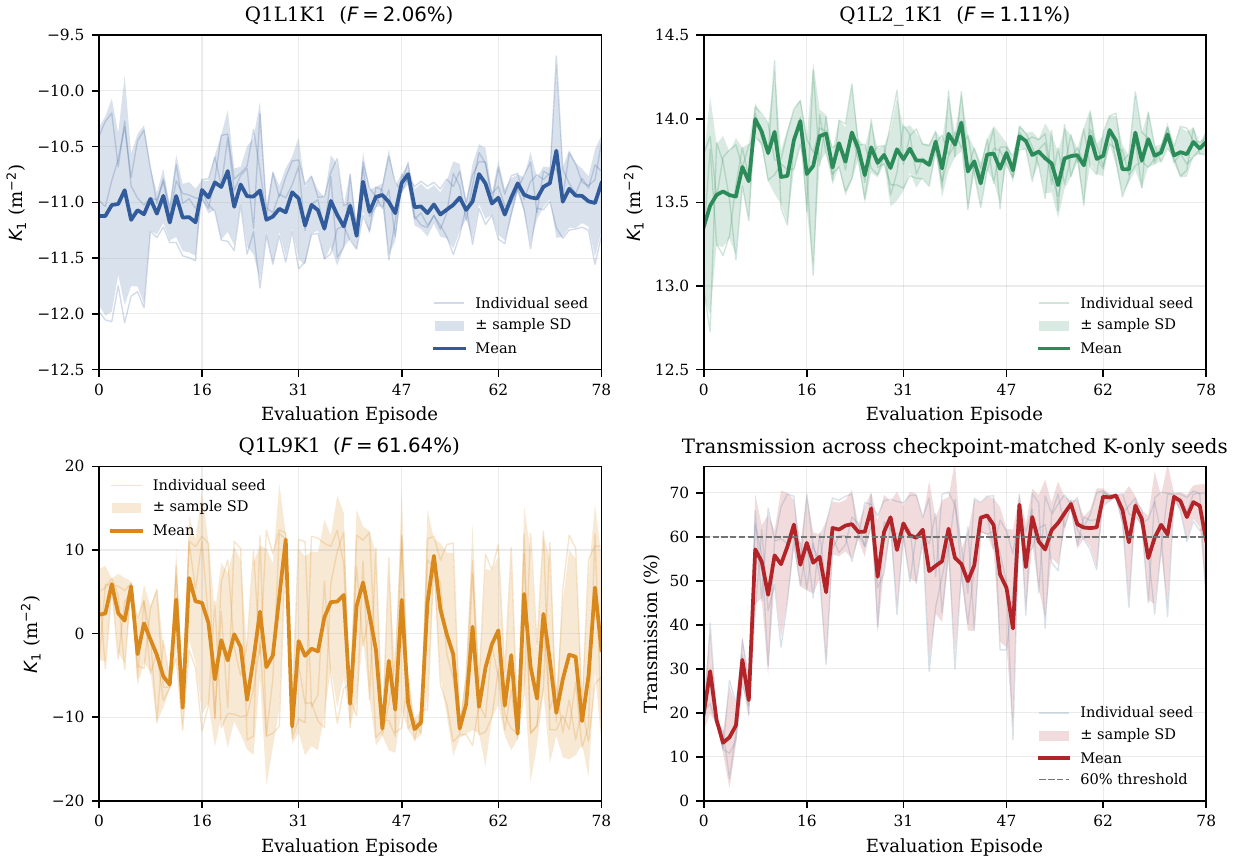}
    \caption{{Representative parameter traces from the evaluation episodes across three independent K-only DDPG runs. Thin lines show the individual seeds, solid lines show the mean, and shaded bands show $\pm$ sample SD. The displayed $F$ values are calculated from all K-only evaluation episodes with transmission above 60\%, with equal total weight for each seed. Q1L1K1 and Q1L2\_1K1 remain within narrow bands, whereas Q1L9K1 varies more broadly.}}
    \label{fig:param_evolution}
\end{figure}

Table~\ref{tab:best_params} presents the optimized parameters from the best-performing configuration (70.3\% transmission). {Figure~\ref{fig:param_evolution} shows representative parameter values across matched evaluation episodes from three independent K-only DDPG runs, illustrating the contrast between narrowly and broadly sampled controls.}

\begin{table}[!ht]
\caption{Optimized quadrupole strengths from best episode. F = focusing ($K_1 > 0$), D = defocusing ($K_1 < 0$).}
\label{tab:best_params}
\centering
\begin{tabular}{lrrll}
\toprule
\textbf{Quadrupole} & \textbf{$K_1$ (m$^{-2}$)} & \textbf{Type} & \textbf{Position} & \textbf{Notes} \\
\midrule
Q1L0 & $+10.17$ & F & Injection & \\
Q1L1 & $-11.09$ & D & Before BM1 & \\
Q1L2\_1 & $+14.07$ & F & After BM1 & \multirow{2}{*}{Symmetric pair} \\
Q1L2\_2 & $+14.33$ & F & After BM3 & \\
Q1L3\_1 & $-5.66$ & D & S-bend middle & \multirow{2}{*}{Symmetric pair} \\
Q1L3\_2 & $-5.39$ & D & S-bend middle & \\
Q1L4\_1 & $+7.47$ & F & S-bend center & \\
Q1L7 & $-12.19$ & D & After BM4 & \\
Q1L4\_2 & $+12.15$ & F & Before debuncher & \\
Q1L9 & $-12.07$ & D & After debuncher & \\
Q1L10 & $+16.72$ & F & Final focusing & \\
\bottomrule
\end{tabular}
\end{table}

\subsection{Optimized Beam Properties}

\begin{figure}[!ht]
\centering
\subfloat[Test beamline: beam envelope ($2\sigma$)]{\includegraphics[width=0.48\textwidth]{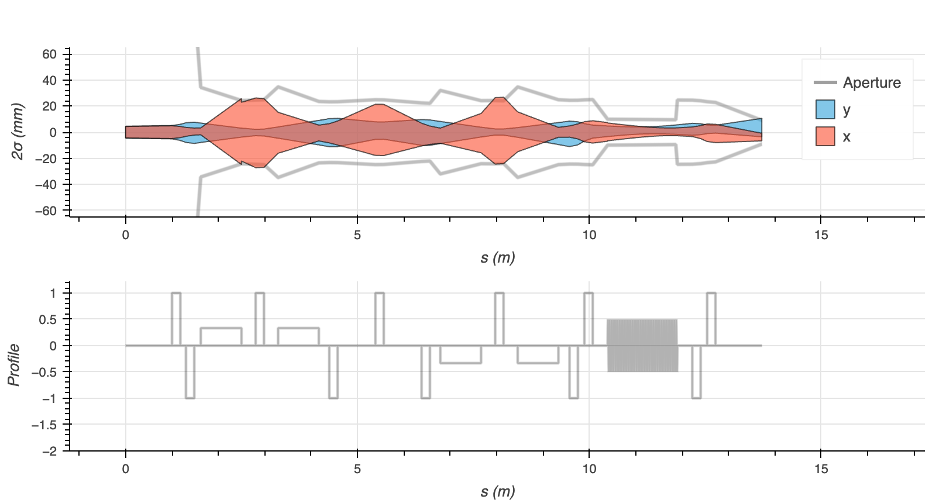}\label{fig:env_vepp5}}
\hfill
\subfloat[Two-dipole: beam envelope ($2\sigma$)]{\includegraphics[width=0.48\textwidth]{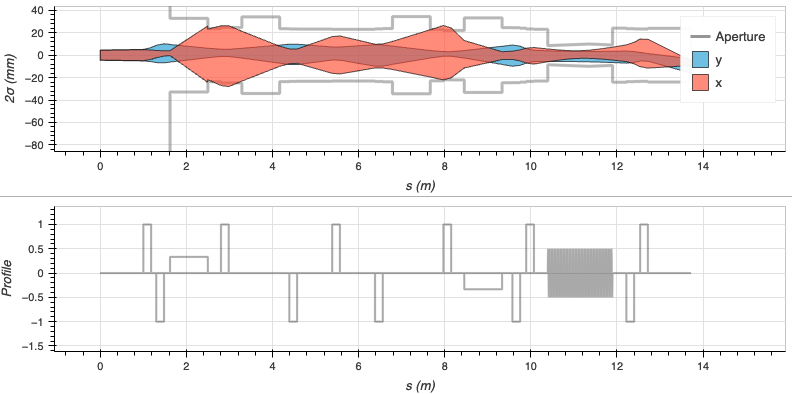}\label{fig:env_2d}}
\\
\subfloat[Test beamline: beta functions]{\includegraphics[width=0.48\textwidth]{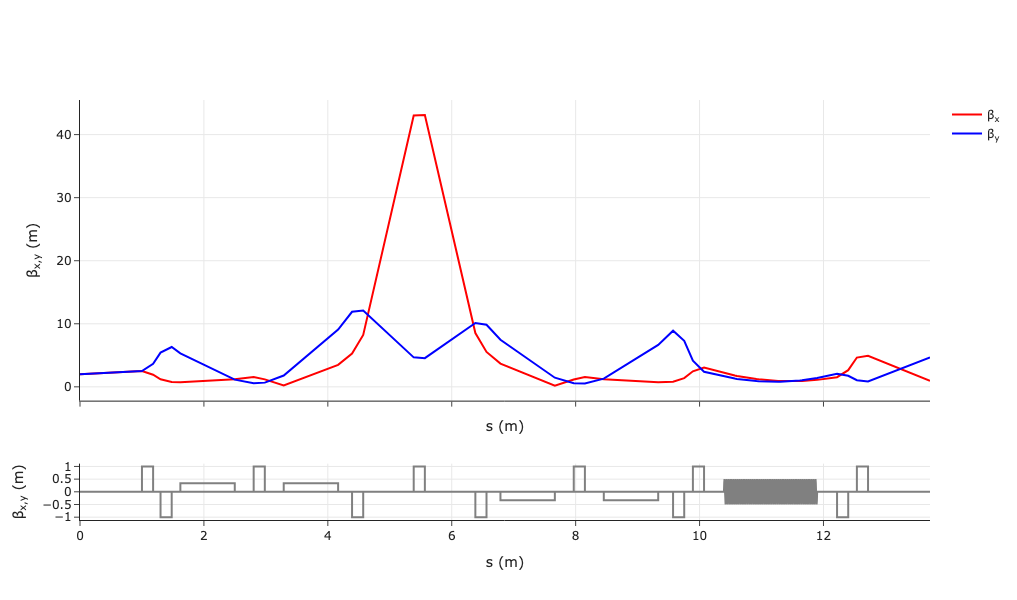}\label{fig:beta_vepp5}}
\hfill
\subfloat[Two-dipole: beta functions]{\includegraphics[width=0.48\textwidth]{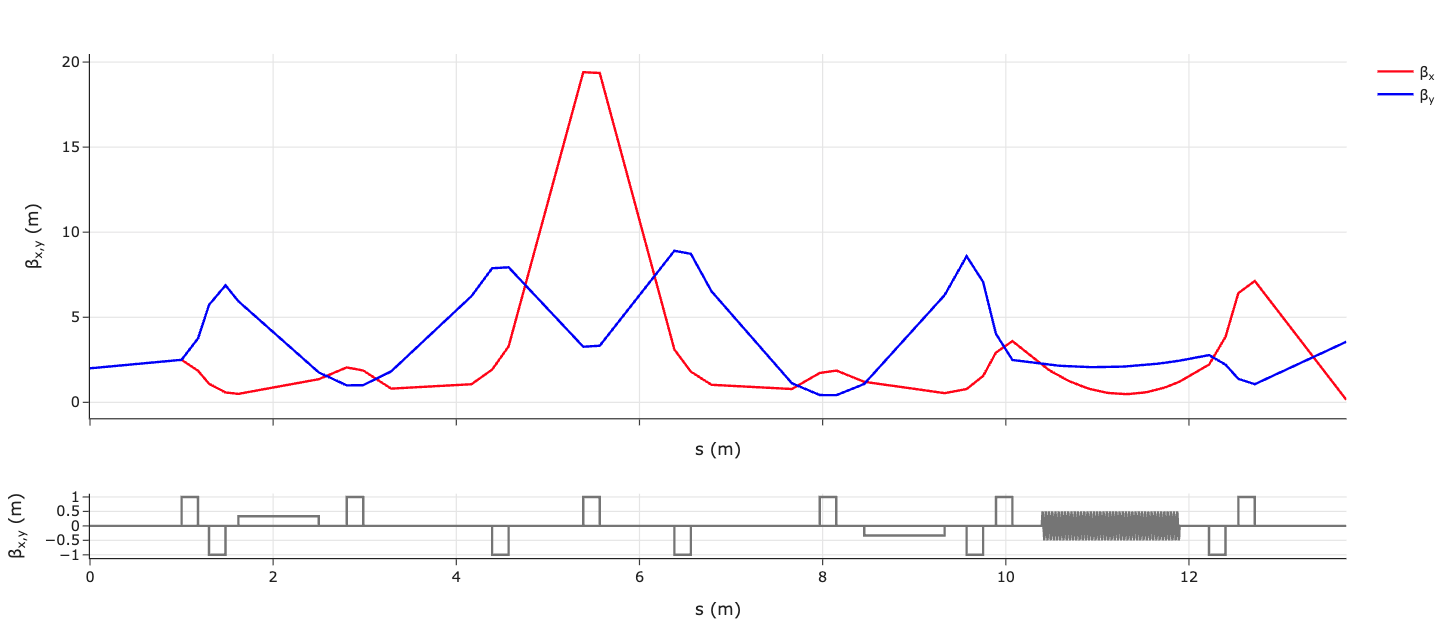}\label{fig:beta_2d}}
\\
\subfloat[Test beamline: dispersion functions]{\includegraphics[width=0.48\textwidth]{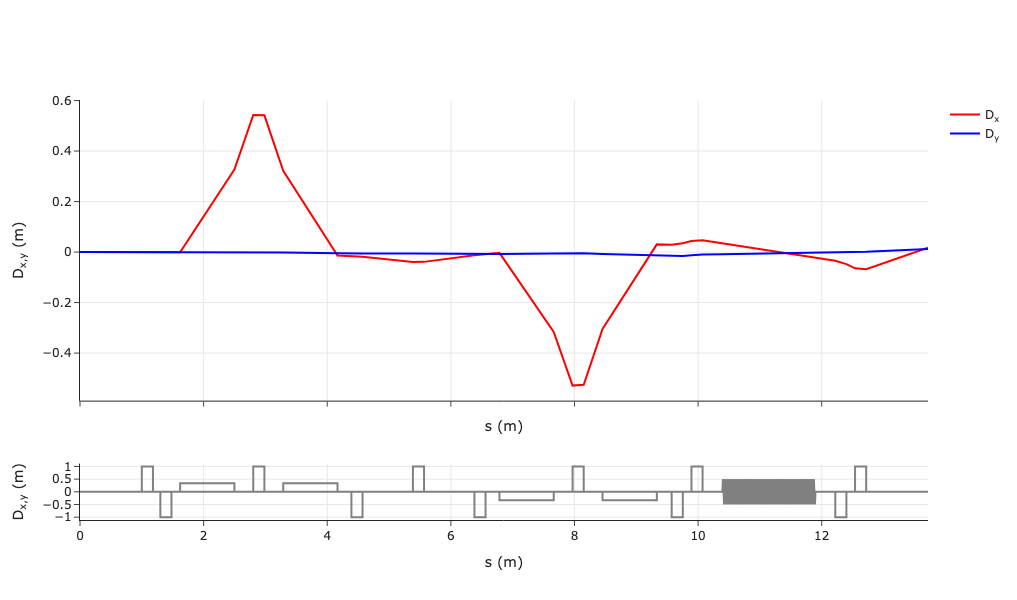}\label{fig:disp_vepp5}}
\hfill
\subfloat[Two-dipole: dispersion functions]{\includegraphics[width=0.48\textwidth]{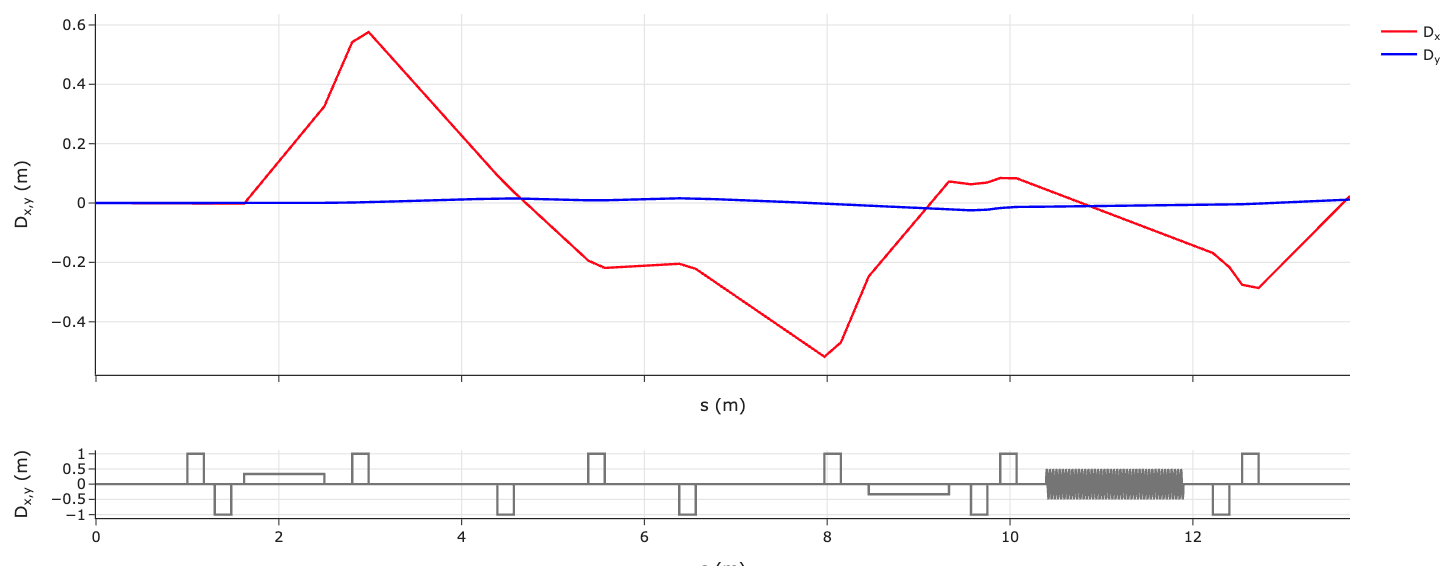}\label{fig:disp_2d}}
\caption{Optimized beam optics for the test beamline (left column) and the two-dipole variant (right column). Top row: beam envelopes ($2\sigma$) remain within aperture limits on both configurations. Middle row: beta functions $\beta_x$ (red) and $\beta_y$ (blue) show anti-correlated alternating-gradient focusing. Bottom row: dispersion functions $D_x$ (red) and $D_y$ (blue); the test beamline's S-bend achieves near-zero dispersion at exit, while the single-bend variant shows a different dispersion profile reflecting its asymmetric geometry.}
\label{fig:beam_optics}
\end{figure}

RLABC's Elegant Wrapper automatically generates standard beam optics plots from the SDDS simulation output, allowing users to quickly assess any optimized configuration. Figure~\ref{fig:beam_optics} presents the beam envelope, Twiss beta functions, and dispersion functions{~\cite{courant1958,wiedemann2015}} for the best configurations found on both the test beamline (left column, 70.3\% transmission) and the two-dipole variant (right column, 70.9\% transmission; see Section~\ref{sec:two_dipole}).

{For the test beamline configuration, the plotted transverse $2\sigma$ envelopes remain within the shown aperture limits, including near the RF debuncher at $s\approx10$--11~m (Figure~\ref{fig:env_vepp5}). The beta functions show the alternating behavior expected from quadrupole focusing (Figure~\ref{fig:beta_vepp5}), and the horizontal dispersion is near zero at the exit in the plotted solution (Figure~\ref{fig:disp_vepp5}). The corresponding plots for the two-dipole variant document its selected transported solution.}

\subsection{{Reuse on a Structurally Different Beamline}}
\label{sec:two_dipole}

\begin{figure}[!htbp]
    \centering
    \includegraphics[width=0.6\textwidth]{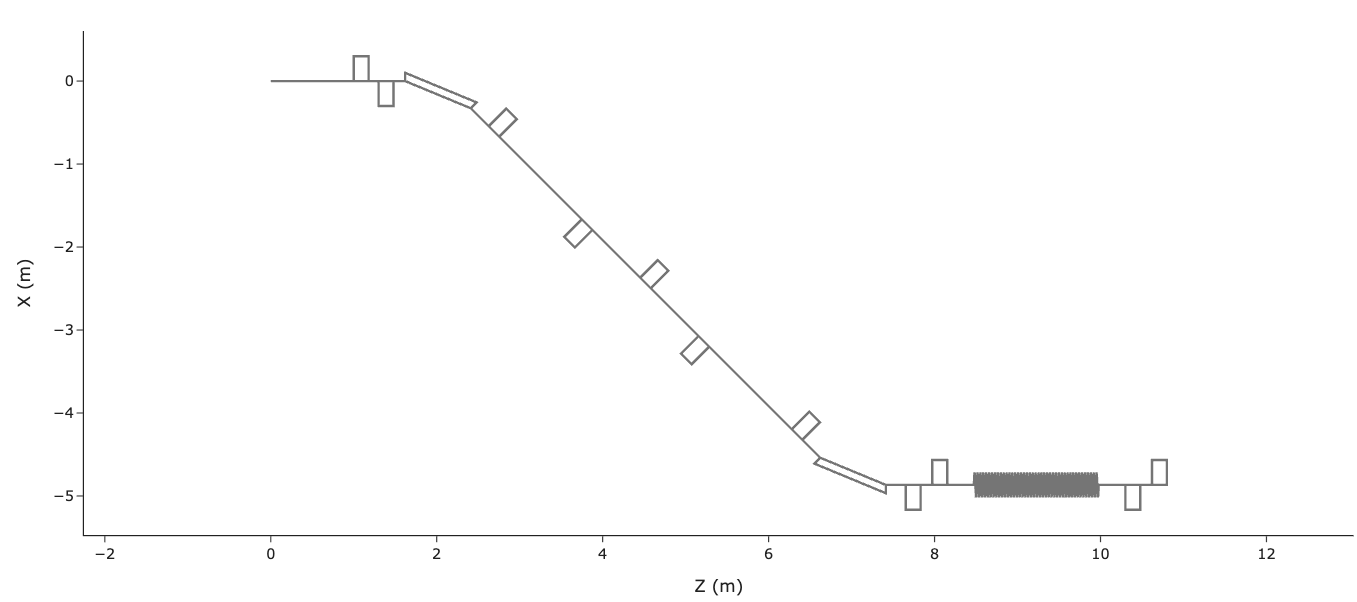}
    \caption{Layout of the two-dipole beamline variant, showing a single-bend geometry with no compensating return bends. The gray block represents the RF cavity.}
    \label{fig:beamline_2dipoles}
\end{figure}

{We also apply the same environment construction to a variant containing two rather than four dipoles (35 tunable parameters), shown in Figure~\ref{fig:beamline_2dipoles}.}

 This configuration produces a single-bend geometry, in contrast to the S-bend topology of the test beamline. The test beamline's S-bend features pairs of opposing dipoles whose bending angles partially compensate each other, yielding a symmetric geometry with naturally self-correcting dispersion. The two-dipole variant lacks this geometric symmetry: dispersion accumulated through the single bending section must be managed entirely by the downstream quadrupole optics, presenting a qualitatively different optimization challenge.

{Using the same observation construction and reward, the DDPG run achieves 70.9\% transmission.}

\subsection{Reproducibility}
\label{sec:reproducibility}

All code, configuration files, and lattice definitions are available at \url{https://github.com/Anwar9Ibrahim/RL-ABC.git} under MIT license. Experiments use:
\begin{itemize}
\item Elegant beam-dynamics simulation and the SDDS Toolkit
\item {Python 3.10}, PyTorch 1.12.1, Stable-Baselines3 1.6.2
\item {Three independent numerical initializations for every experiment in Table~\ref{tab:multiseed_validation}, with the run-level outputs retained in the evaluation archive; the parameter-spread analysis uses the selected DDPG configurations from these runs.}
\item {The full hyperparameter configuration is available in \path{config.json} in the GitHub repository and is summarized in Table~\ref{tab:hyperparams}}
\end{itemize}

{The DDPG training hyperparameters are listed in Table~\ref{tab:hyperparams}. Each K-only run and each all-variable run trained from scratch consists of 30,000 episodes. For both experiments, the Gaussian exploration-noise scale decays linearly from 0.5 to 0.05 during the first 10,000 episodes and is then held at 0.05 for another 20,000 episodes. The warm-start experiment uses the policy obtained after the first 10,000 episodes of K-only training, then continues with all variables active for 10,000 episodes while the exploration-noise scale decays linearly from 0.2 to 0.05, followed by another 10,000 episodes at 0.05. Thus, including the K-only pretraining, the warm-start pipeline also comprises 30,000 episodes. Exploration noise is added to the normalized action and the perturbed action is clipped to $[-1,1]^4$; the periodic evaluation episodes are noise-free. Neither the actor nor the critic learning rate is annealed.}

\begin{table}[!ht]
\caption{{DDPG training hyperparameters used by the reported configuration.}}
\label{tab:hyperparams}
\centering
{%
\begin{tabular}{@{}p{0.40\textwidth}p{0.54\textwidth}@{}}
\toprule
\textbf{Hyperparameter} & \textbf{Value} \\
\midrule
Discount factor $\gamma$ & $0.99$ \\
Actor / critic learning rate (Adam) & $10^{-4}$ / $10^{-3}$ \\
Batch size & $128$ \\
Replay buffer capacity & $10^{6}$ \\
{Replay transition masks} & {Current and next 4D masks stored per transition} \\
Target soft-update rate $\tau$ & $0.005$ \\
Learning-rate schedule & None; both Adam rates held constant \\
Exploration noise & Zero-mean Gaussian added in the normalized action space, then clipped; per-experiment scales below \\
Training episodes per run & $30{,}000$ (K-only); $30{,}000$ (all variables from scratch); $10{,}000$ K-only pretraining $+20{,}000$ all-variable continuation (warm start) \\
Exploration noise scale & K-only and from scratch: $0.5\rightarrow0.05$ over the first $10{,}000$ episodes, then $0.05$ for $20{,}000$; warm-start continuation: $0.2\rightarrow0.05$ over $10{,}000$, then $0.05$ for $10{,}000$ \\
Normalized / physical action bounds & $[-1,1]^4$ / Table~\ref{tab:actions} \\
Hidden layers (actor \& critic) & $800, 600, 512, 256$ (layer norm) \\
Histogram bins $n_{\text{bins}}$ & $5$ ($5\times5$) \\
\bottomrule
\end{tabular}%
}
\end{table}

{For the classical comparisons, differential evolution uses a population multiplier of 15, Latin-hypercube initialization, mutation factor 0.8, crossover probability 0.9, and no polishing. This gives 165 candidates and 33,000 objective evaluations per K-only run, and 555 candidates and 33,300 objective evaluations per full-variable run, with $N_0=10^3$ particles per evaluation. The independently initialized Bayesian-optimization experiments use 10 initial design points followed by an upper-confidence-bound acquisition with $\kappa=2.576$ and 1,000 objective evaluations per run. The K-only Bayesian-optimization search uses $N_0=10^5$ particles per evaluation, while the all-variable search uses $N_0=10^3$. The separate informed-initial-design run uses 500 evaluations and a predefined initial design containing the high-transmission configuration discussed above.}

\clearpage
\section{Conclusion}
\label{sec:conclusion}

We present \textbf{RLABC}, an open-source Python framework that automates the transformation of {Elegant-enabled} beamline configurations into reinforcement learning environments. The primary contribution is methodological: a general approach for formulating beamline tuning as a Markov decision process through automatic lattice preprocessing, a fixed 57-dimensional {observation representation}, and integration with the widely-used Elegant simulation code via SDDS interfaces.

Validation on a test beamline derived from the VEPP-5 injection complex demonstrates that RLABC successfully enables RL-based optimization. {Across three K-only runs, the selected policies have a mean transmission of 69.81\% and a sample SD of 0.73 percentage points when each selected configuration is re-evaluated with $N_0=10^5$ particles. A short K-only warm start improves early full-variable adaptation, while its selected-policy endpoint is similar to training from scratch.} Additional validation on a structurally different two-dipole beamline variant - featuring an asymmetric single-bend geometry in contrast to the test beamline's S-bend - yields 70.9\% transmission (35 parameters).

We expect RLABC to be useful both for accelerator physicists who want to experiment with RL-based tuning without building environments from scratch - only standard Elegant lattice files are needed as input - and for RL researchers looking for a physically grounded benchmark with continuous actions, nonlinear dynamics, and meaningful constraints.

A current limitation is the high computational cost of RL training relative to classical optimizers, inherent to the sample complexity of deep RL in high-dimensional continuous spaces. {Because the present paper focuses primarily on the RLABC tool and its environment-construction methodology, future work will investigate the agent component more systematically. This will include matched-budget, multi-seed comparisons of alternative continuous-control algorithms such as SAC, PPO, and TD3; evaluation of different stage-learning curricula; tests on larger and more complex beamlines; and controlled studies of cross-beamline transfer learning. Additional work will explore extension to real accelerator deployment.}

\section*{Acknowledgments}

The authors thank Alexander Barnyakov and the VEPP-5 operations team at BINP for providing facility data and operational insights. The computation for this research was performed using the computational resources of HPC facilities at HSE University.

\section*{CRediT authorship contribution statement}

\textbf{Anwar Ibrahim:} Conceptualization, Methodology, Software, Validation, Formal analysis, Investigation, Data curation, Writing -- original draft, Visualization. \textbf{Fedor Ratnikov:} Conceptualization, Methodology, Supervision, Writing -- review \& editing. \textbf{Maxim Kaledin:} Conceptualization, Software, Validation, Writing -- review \& editing. \textbf{Alexey Petrenko:} Conceptualization, Methodology, Validation, Resources, Data curation, Writing -- review \& editing, Supervision. \textbf{Denis Derkach:} Conceptualization, Methodology, Supervision, Funding acquisition, Project administration, Writing -- review \& editing.

\section*{Declaration of competing interest}

The authors declare that they have no known competing financial interests or personal relationships that could have appeared to influence the work reported in this paper.

\section*{Data availability}

The source code, configuration files, and example lattice files are publicly available at \url{https://github.com/Anwar9Ibrahim/RL-ABC.git} under the MIT license. The test beamline lattice configuration used in this study is included in the repository.

\section*{Funding}

The work is supported by the grant for research centers in the field of AI provided by the Ministry of Economic Development of the Russian Federation in accordance with the agreement 000000C313925P4E0002 and the agreement with HSE University № 139-15-2025-009.

\section*{Declaration of generative AI and AI-assisted technologies in the writing process}

During the preparation of this work the authors used ChatGPT (OpenAI) and Claude (Anthropic) in order to improve the language and readability of the manuscript. After using these tools, the authors reviewed and edited the content as needed and take full responsibility for the content of the published article.


\bibliographystyle{elsarticle-num}
\bibliography{main}

@book{lee2018,
  title={Accelerator physics},
  author={Lee, Shyh-Yuan},
  year={2018},
  publisher={World Scientific Publishing Company}
}

@book{chao2013,
    editor = "Chao, Alexander Wu and Mess, Karl Hubert and Tigner, Maury and Zimmermann, Frank",
    title = "{Handbook of accelerator physics and engineering}: {2nd Edition}",
    edition = "2",
    doi = "10.1142/8543",
    isbn = "978-981-4415-84-2",
    publisher = "World Scientific",
    address = "Hackensack, USA",
    year = "2013"
}

@book{sutton2018,
author = {Sutton, Richard S. and Barto, Andrew G.},
title = {Reinforcement Learning: An Introduction},
year = {2018},
isbn = {0262039249},
publisher = {A Bradford Book},
address = {Cambridge, MA, USA}
}

@ARTICLE{mnih2015,
  title    = "Human-level control through deep reinforcement learning",
  author   = "Mnih, Volodymyr and Kavukcuoglu, Koray and Silver, David and
              Rusu, Andrei A and Veness, Joel and Bellemare, Marc G and Graves,
              Alex and Riedmiller, Martin and Fidjeland, Andreas K and
              Ostrovski, Georg and Petersen, Stig and Beattie, Charles and
              Sadik, Amir and Antonoglou, Ioannis and King, Helen and Kumaran,
              Dharshan and Wierstra, Daan and Legg, Shane and Hassabis, Demis",
  journal  = "Nature",
  volume   =  518,
  number   =  7540,
  pages    = "529--533",
  month    =  feb,
  year     =  2015
}

@article{kober2013,
  title={Reinforcement learning in robotics: A survey},
  author={Kober, Jens and Bagnell, J Andrew and Peters, Jan},
  journal={The International Journal of Robotics Research},
  volume={32},
  number={11},
  pages={1238--1274},
  year={2013},
  publisher={SAGE Publications Sage UK: London, England}
}

@techreport{borland2000,
  title={Elegant: A flexible SDDS-compliant code for accelerator simulation},
  author={Borland, Michael},
  year={2000},
  institution={Argonne National Lab., IL (US)}
}

@book{wiedemann2015,
  title={Particle accelerator physics},
  author={Wiedemann, Helmut},
  year={2015},
  publisher={Springer Nature}
}

@book{wolski2014,
  title={Beam dynamics in high energy particle accelerators},
  author={Wolski, Andrzej},
  year={2014},
  publisher={World Scientific}
}

@article{nelder1965,
  title={A simplex method for function minimization},
  author={Nelder, John A and Mead, Roger},
  journal={The computer journal},
  volume={7},
  number={4},
  pages={308--313},
  year={1965},
  publisher={The British Computer Society}
}

@article{raffin2021,
  title={Stable-baselines3: Reliable reinforcement learning implementations},
  author={Raffin, Antonin and Hill, Ashley and Gleave, Adam and Kanervisto, Anssi and Ernestus, Maximilian and Dormann, Noah},
  journal={Journal of machine learning research},
  volume={22},
  number={268},
  pages={1--8},
  year={2021}
}

@inproceedings{kendall2018learning,
  title={Learning to drive in a day},
  author={Kendall, Alex and Hawke, Jeffrey and Janz, David and Mazur, Przemyslaw and Reda, Daniele and Allen, John-Mark and Lam, Vinh-Dieu and Bewley, Alex and Shah, Amar},
  booktitle={2019 international conference on robotics and automation (ICRA)},
  pages={8248--8254},
  year={2019},
  organization={IEEE}
}

@inproceedings{tang2025deep,
  title={Deep reinforcement learning for robotics: A survey of real-world successes},
  author={Tang, Chen and Abbatematteo, Ben and Hu, Jiaheng and Chandra, Rohan and Mart{\'\i}n-Mart{\'\i}n, Roberto and Stone, Peter},
  booktitle={Proceedings of the AAAI Conference on Artificial Intelligence},
  volume={39},
  pages={28694--28698},
  year={2025}
}

@article{bai2025review,
  title={A review of reinforcement learning in financial applications},
  author={Bai, Yahui and Gao, Yuhe and Wan, Runzhe and Zhang, Sheng and Song, Rui},
  journal={Annual Review of Statistics and Its Application},
  volume={12},
  number={1},
  pages={209--232},
  year={2025},
  publisher={Annual Reviews}
}

@inproceedings{nambiar2023deep,
  title={Deep offline reinforcement learning for real-world treatment optimization applications},
  author={Nambiar, Mila and Ghosh, Supriyo and Ong, Priscilla and Chan, Yu En and Bee, Yong Mong and Krishnaswamy, Pavitra},
  booktitle={Proceedings of the 29th ACM SIGKDD conference on knowledge discovery and data mining},
  pages={4673--4684},
  year={2023}
}

@ARTICLE{Ibrahim2025PPN,
  title    = "Optimisation of the Accelerator Control by Reinforcement
              Learning: A {Simulation-Based} Approach",
  author   = "Ibrahim, Anwar and Derkach, Denis and Petrenko, Alexey and Ratnikov, Fedor and Kaledin, Maxim",
  journal  = "Physics of Particles and Nuclei",
  volume   =  56,
  number   =  6,
  pages    = "1476--1481",
  month    =  dec,
  year     =  2025,
  doi      = "10.1134/S1063779625700716"
}

@article{ibrahim2025IJMP,
  title   = {Reinforcement Learning for Accelerator Beamline Control: a simulation-based approach},
  author  = {Ibrahim, Anwar and Petrenko, Alexey and Kaledin, Maxim and Suleiman, Ehab and Ratnikov, Fedor and Derkach, Denis},
  journal = {International Journal of Modern Physics E},
  year    = {2026},
  pages   = {2641027},
  doi     = {10.1142/S0218301326410272},
  note    = {arXiv:2510.26805}
}

@article{kaiser2024bridging,
  author = {Kaiser, Jan and Xu, Chenran and Eichler, Annika and Santamaria Garcia, Andrea},
  title = {Bridging the gap between machine learning and particle accelerator physics with high-speed, differentiable simulations},
  journal = {Physical Review Accelerators and Beams},
  volume = {27},
  number = {5},
  pages = {054601},
  year = {2024},
  doi = {10.1103/PhysRevAccelBeams.27.054601},
  url = {https://doi.org/10.1103/PhysRevAccelBeams.27.054601},
}

@article{scomparin2024microsecond,
  author = {Scomparin, Luca and Caselle, Michele and Santamaria Garcia, Andrea and Xu, Chenran and Blomley, Edmund and Dritschler, Timo and Mochihashi, Akira and Schuh, Marcel and Steinmann, Johannes L. and Bründermann, Erik and Kopmann, Andreas and Becker, Jürgen and Müller, Anke-Susanne and Weber, Marc},
  title = {Microsecond-Latency Feedback at a Particle Accelerator by Online Reinforcement Learning on Hardware},
  journal = {arXiv preprint arXiv:2409.16177},
  year = {2024},
  url = {https://arxiv.org/abs/2409.16177},
}

@article{meier2022srf,
  title   = {Optimizing a superconducting radiofrequency gun using deep reinforcement learning},
  author  = {Meier, David and Ramirez, Luis Vera and Völker, Jens and Sick, Bernhard and Viefhaus, Jens and Hartmann, Gregor},
  journal = {Physical Review Accelerators and Beams},
  volume  = {25},
  number  = {10},
  pages   = {104604},
  year    = {2022},
  doi     = {10.1103/PhysRevAccelBeams.25.104604}
}

@article{chen2023trend,
  title         = {Trend-Based {SAC} Beam Control Method with Zero-Shot in Superconducting Linear Accelerator},
  author        = {Chen, Xiaolong and Qi, Xin and Su, Chunguang and He, Yuan and Wang, Zhijun and Sun, Kunxiang and Jin, Chao and Chen, Weilong and Liu, Shuhui and Zhao, Xiaoying and Jia, Duanyang and Yi, Man},
  journal       = {arXiv preprint arXiv:2305.13869},
  year          = {2023},
  doi           = {10.48550/arXiv.2305.13869},
  archivePrefix = {arXiv},
  eprint        = {2305.13869}
}

@article{rousseletal2024,
  title        = {Bayesian optimization algorithms for accelerator physics},
  author       = {Roussel, Ryan and Edelen, Auralee L. and Boltz, Tobias and Kennedy, Dylan and Zhang, Zhe and Ji, Fuhao and Huang, Xiaobiao and Ratner, Daniel and Santamaria Garcia, Andrea and Xu, Chenran and Kaiser, Jan and Ferran Pousa, Angel and Eichler, Annika and Lubsen, Jannis O. and Isenberg, Natalie M. and Gao, Yuan and Kuklev, Nikita and Martinez, Jose and Mustapha, Brahim and Kain, Verena and Lin, Weijian and Liuzzo, Simone Maria and St. John, Jason and Streeter, Matthew J. V. and Lehe, Remi and Neiswanger, Willie},
  journal      = {Physical Review Accelerators and Beams},
  volume       = {27},
  number       = {8},
  pages        = {084801},
  year         = {2024},
  doi          = {10.1103/PhysRevAccelBeams.27.084801},
  publisher    = {American Physical Society},
  note         = {Review of Bayesian optimization methods in accelerator physics} 
}

@article{hanuka2021,
  title        = {Physics model-informed Gaussian process for online optimization of particle accelerators},
  author       = {Hanuka, Adi and Huang, X. and Shtalenkova, J. and Kennedy, D. and Edelen, A. and Zhang, Z. and Lalchand, V. R. and Ratner, D. and Duris, J.},
  journal      = {Physical Review Accelerators and Beams},
  volume       = {24},
  number       = {7},
  pages        = {072802},
  year         = {2021},
  doi          = {10.1103/PhysRevAccelBeams.24.072802},
  publisher    = {American Physical Society},
  note         = {Physics-informed prior mean and kernel improve convergence of Gaussian-process optimization}
}

@article{morita2023,
  title        = {Accelerator tuning method using autoencoder and Bayesian optimization},
  author       = {Morita, Yasuyuki and Washio, Takashi and Nakashima, Yuta},
  journal      = {Nuclear Instruments and Methods in Physics Research Section A: Accelerators, Spectrometers, Detectors and Associated Equipment},
  volume       = {1057},
  pages        = {168730},
  year         = {2023},
  doi          = {10.1016/j.nima.2023.168730},
  note         = {Autoencoder-assisted Bayesian optimization for high-dimensional accelerator tuning} 
}

@article{awal2025,
  title        = {Injection optimization at particle accelerators via reinforcement learning: From simulation to real-world application},
  author       = {Awal, Awal and Hetzel, Jan and Gebel, Ralf and Pretz, Jörg},
  journal      = {Physical Review Accelerators and Beams},
  volume       = {28},
  number       = {3},
  pages        = {034601},
  year         = {2025},
  doi          = {10.1103/PhysRevAccelBeams.28.034601},
  note         = {RL framework for optimizing the injection process with SAC and live evaluation} 
}

@article{kaiser2024,
  title        = {Reinforcement learning-trained optimisers and Bayesian optimisation for online particle accelerator tuning},
  author       = {Kaiser, Jan and Xu, Chenran and Eichler, Annika and Santamaria Garcia, Andrea and Stein, Oliver and Br{\"u}ndermann, Erik and Kuropka, Willi and Dinter, Hannes and Mayet, Frank and Vinatier, Thomas and Burkart, Florian and Schlarb, Holger},
  journal      = {Scientific Reports},
  volume       = {14},
  pages        = {15733},
  year         = {2024},
  doi          = {10.1038/s41598-024-66263-y},
  note         = {Comparative study of RL and Bayesian optimization for online tuning} 
}

@misc{2025gymnasiums,
      title={Gymnasium: A Standard Interface for Reinforcement Learning Environments}, 
      author={Mark Towers and Ariel Kwiatkowski and Jordan Terry and John U. Balis and Gianluca De Cola and Tristan Deleu and Manuel Goulão and Andreas Kallinteris and Markus Krimmel and Arjun KG and Rodrigo Perez-Vicente and Andrea Pierré and Sander Schulhoff and Jun Jet Tai and Hannah Tan and Omar G. Younis},
      year={2025},
      eprint={2407.17032},
      archivePrefix={arXiv},
      primaryClass={cs.LG},
      url={https://arxiv.org/abs/2407.17032}, 
}

@inproceedings{silver:hal-00938992,
  TITLE = {{Deterministic Policy Gradient Algorithms}},
  AUTHOR = {Silver, David and Lever, Guy and Heess, Nicolas and Degris, Thomas and Wierstra, Daan and Riedmiller, Martin},
  URL = {https://inria.hal.science/hal-00938992},
  BOOKTITLE = {{ICML}},
  PAGES = {387--395},
  ADDRESS = {Beijing, China},
  YEAR = {2014},
  MONTH = Jun,
  HAL_ID = {hal-00938992},
  HAL_VERSION = {v1},
}

@ARTICLE{2020SciPy-NMeth,
  author  = {Virtanen, Pauli and Gommers, Ralf and Oliphant, Travis E. and
            Haberland, Matt and Reddy, Tyler and Cournapeau, David and
            Burovski, Evgeni and Peterson, Pearu and Weckesser, Warren and
            Bright, Jonathan and {van der Walt}, St{\'e}fan J. and
            Brett, Matthew and Wilson, Joshua and Millman, K. Jarrod and
            Mayorov, Nikolay and Nelson, Andrew R. J. and Jones, Eric and
            Kern, Robert and Larson, Eric and Carey, C J and
            Polat, {\.I}lhan and Feng, Yu and Moore, Eric W. and
            {VanderPlas}, Jake and Laxalde, Denis and Perktold, Josef and
            Cimrman, Robert and Henriksen, Ian and Quintero, E. A. and
            Harris, Charles R. and Archibald, Anne M. and
            Ribeiro, Ant{\^o}nio H. and Pedregosa, Fabian and
            {van Mulbregt}, Paul and {SciPy 1.0 Contributors}},
  title   = {{{SciPy} 1.0: Fundamental Algorithms for Scientific
            Computing in Python}},
  journal = {Nature Methods},
  year    = {2020},
  volume  = {17},
  pages   = {261--272},
  adsurl  = {https://rdcu.be/b08Wh},
  doi     = {10.1038/s41592-019-0686-2},
}

@incollection{pelikan2005bayesian,
  title={Bayesian optimization algorithm},
  author={Pelikan, Martin},
  booktitle={Hierarchical Bayesian optimization algorithm: toward a new generation of evolutionary algorithms},
  pages={31--48},
  year={2005},
  publisher={Springer}
}

@misc{Nogueira2014BayesianOptimization,
  author = {Nogueira, Fernando},
  title = {Bayesian Optimization: Open Source Constrained Global Optimization Tool for Python},
  year = {2014--},
  url = {https://github.com/bayesian-optimization/BayesianOptimization}
}

@ARTICLE{Emanov2023-wx,
  title    = "{VEPP-5} Injection Complex",
  author   = "Emanov, F A and Astrelina, K V and Balakin, V V and Belikov, O V
              and Berkaev, D E and Boimelshtain, Yu M and Bolkhovityanov, D Yu
              and Frolov, A R and Karpov, G V and Kasaev, A S and Kondakov, A A
              and Kurkin, G Ya and Lapik, R M and Lebedev, N N and Levichev, A
              E and Maltseva, Yu I and Murasev, A A and Samoylov, S L and
              Vasiliev, S V and Martinovsky, A Yu and Motygin, C V and Pilan, A
              M and Tribendis, A G and Pavlenko, A V and Kotov, E S and
              Arsentyeva, M V",
  journal  = "Physics of Particles and Nuclei Letters",
  volume   =  20,
  number   =  4,
  pages    = "754--756",
  month    =  aug,
  year     =  2023,
  doi      = "10.1134/S1547477123040258"
}

@ARTICLE{Storn1997-db,
  title    = "Differential Evolution -- A Simple and Efficient Heuristic for
              global Optimization over Continuous Spaces",
  author   = "Storn, Rainer and Price, Kenneth",
  journal  = "Journal of Global Optimization",
  volume   =  11,
  number   =  4,
  pages    = "341--359",
  month    =  dec,
  year     =  1997
}

@inproceedings{grote2003madx,
  author       = {Grote, H. and Schmidt, F.},
  title        = {{MAD-X}: An Upgrade from {MAD8}},
  booktitle    = {Proceedings of the 2003 Particle Accelerator Conference},
  series       = {PAC 2003},
  pages        = {3497--3499},
  year         = {2003},
  organization = {IEEE},
  address      = {Portland, OR, USA},
  note         = {CERN-AB-2003-024-ABP}
}

@manual{floettmann2017astra,
  author       = {Floettmann, Klaus},
  title        = {{ASTRA}: A Space Charge Tracking Algorithm, User's Manual},
  organization = {DESY},
  address      = {Hamburg, Germany},
  edition      = {Version 3.2},
  month        = {March},
  year         = {2017},
  url          = {https://www.desy.de/~mpyflo/},
  note         = {DESY, Notkestr. 85, 22603 Hamburg, Germany}
}

@article{lillicrap2015,
  title={Continuous control with deep reinforcement learning},
  author={Lillicrap, Timothy P and Hunt, Jonathan J and Pritzel, Alexander and Heess, Nicolas and Erez, Tom and Tassa, Yuval and Silver, David and Wierstra, Daan},
  journal={arXiv preprint arXiv:1509.02971},
  year={2015},
  note={Presented at ICLR 2016}
}

@article{Kain2020,
  title   = {Sample-efficient reinforcement learning for {CERN} accelerator control},
  author  = {Kain, Verena and Hirlaender, Simon and Goddard, Brennan and Velotti, Francesco Maria and Zevi Della Porta, Giovanni and Bruchon, Niky and Valentino, Gianluca},
  journal = {Physical Review Accelerators and Beams},
  volume  = {23},
  number  = {12},
  pages   = {124801},
  year    = {2020},
  doi     = {10.1103/PhysRevAccelBeams.23.124801}
}

@article{StJohn2021,
  title   = {Real-time artificial intelligence for accelerator control: A study at the {Fermilab} {Booster}},
  author  = {St. John, Jason and Herwig, Christian and Kafkes, Diana and Mitrevski, Jovan and Pellico, William A. and Perdue, Gabriel N. and Quintero-Parra, Andres and Schupbach, Brian A. and Seiya, Kiyomi and Tran, Nhan and Schram, Malachi and Duarte, Javier M. and Huang, Yunzhi and Keller, Rachael},
  journal = {Physical Review Accelerators and Beams},
  volume  = {24},
  number  = {10},
  pages   = {104601},
  year    = {2021},
  doi     = {10.1103/PhysRevAccelBeams.24.104601}
}

@article{velotti2023,
  title   = {Towards automatic setup of 18 {MeV} electron beamline using machine learning},
  author  = {Velotti, Francesco Maria and Goddard, Brennan and Kain, Verena and Ramjiawan, Rebecca and Zevi Della Porta, Giovanni and Hirlaender, Simon},
  journal = {Machine Learning: Science and Technology},
  volume  = {4},
  number  = {2},
  pages   = {025016},
  year    = {2023},
  doi     = {10.1088/2632-2153/acce21}
}

@article{courant1958,
  title={Theory of the alternating-gradient synchrotron},
  author={Courant, Ernest D and Snyder, Hartland S},
  journal={Annals of physics},
  volume={3},
  number={1},
  pages={1--48},
  year={1958},
  publisher={Elsevier},
  doi={10.1016/0003-4916(58)90012-5}
}

@inproceedings{mckinney2010pandas,
  author={McKinney, Wes},
  title={Data Structures for Statistical Computing in Python},
  booktitle={Proceedings of the 9th Python in Science Conference},
  editor={van der Walt, St\'efan and Millman, Jarrod},
  pages={56--61},
  year={2010},
  doi={10.25080/Majora-92bf1922-00a}
}

\end{document}